\documentclass[10pt,journal,final]{IEEEtran}
\IEEEoverridecommandlockouts

\usepackage{amsfonts,amssymb}
\usepackage{ifpdf}
\usepackage{cite}
\usepackage{stfloats}
\usepackage{bm}
\usepackage{color,xcolor}
\usepackage[caption=false,font=footnotesize]{subfig}

\usepackage{booktabs}      
\usepackage{multirow}      
\usepackage{makecell}      
\usepackage{tabularx}      
\usepackage{threeparttable}
\usepackage{siunitx}       
\ifCLASSINFOpdf
\usepackage[pdftex]{graphicx}
\graphicspath{{../pdf/}{../jpeg/}}
\DeclareGraphicsExtensions{.pdf,.jpeg,.png}
\else
\usepackage[dvips]{graphicx}

\graphicspath{{../eps/}}

\DeclareGraphicsExtensions{.eps}
\fi

\usepackage[cmex10]{amsmath}

\usepackage{algorithm}
\usepackage{algorithmic}
\ifCLASSOPTIONcompsoc
\else
\usepackage[caption=false,font=footnotesize]{subfig}
\fi

\usepackage{makecell}

\begin{document}

\title{Resolving Multi-Target Association in OFDM-based ISAC via Vision-aided Multi-Modal Learning}
\author{
	Meng Hua,~\IEEEmembership{Senior Member,~IEEE,}
	Chenghong~Bian,~\IEEEmembership{Member,~IEEE},
	and Deniz~G\"und\"uz,~\IEEEmembership{Fellow,~IEEE}
		\thanks{This work was  supported by the SNS JU Project 6G-GOALS under the	EU’s Horizon Program with Grant  101139232.
	}

	\thanks{M. Hua and D. G\"und\"uz are with the Department of Electrical and Electronic Engineering, Imperial College London, London SW7 2AZ, U.K. (e-mail: \{m.hua, d.gunduz\}@imperial.ac.uk); C. Bian is with the Department of Electrical and Computer
		Engineering, Hong Kong University of Science and Technology, Hong	Kong. (e-mail: eechbian@ust.hk).
	}

}
\maketitle
\vspace{-1.2cm}
\begin{abstract}

Orthogonal frequency division multiplexing (OFDM)-based  integrated sensing and communication (ISAC)  systems commonly extract target parameters by peak-searching a delay-Doppler map (DDM) constructed from reflected pilots. In multi-target scenarios, this results in ambiguity: the DDM does not reveal which physical target produced which peak, and two targets within the same delay-Doppler resolution cell cannot be separated. We propose a vision-assisted OFDM-ISAC framework that resolves both limitations by fusing wireless and visual modalities. The transmitter encodes an onboard street-view image with deep joint source-channel coding (DeepJSCC) and transmits it over the same OFDM waveform used for sensing; the receiver reconstructs the image, runs a fine-tuned YOLOv5 detector and fuses the resulting per-target features (bounding-box coordinates and class labels) with the DDM and transmitter-receiver geometry through a learned multi-modal network. To stabilize training of the high-dimensional delay/Doppler classifiers, we introduce a Kullback–Leibler loss against triangular soft labels centered on the ground-truth bin. On a Blender-rendered vehicular testbed, the proposed framework achieves a  16 cm localization root mean square error (RMSE), a 10.8 ns   delay RMSE, and a 5.5 m/s  velocity RMSE. An ablation study confirms that removing the visual modality causes a 60$\times$ degradation in localization. These results highlight the potential of vision to overcome the  data-association and resolution limits of single-modality ISAC.
\end{abstract}

\section{Introduction}
Integrated sensing and communication (ISAC) has recently emerged as a key use case  for the sixth-generation (6G) of mobile networks, where  spectrum and hardware resources for sensing and communication are integrated into a unified platform \cite{ Liu2022survey}.   By  leveraging advanced signal processing and machine learning techniques, ISAC is envisioned to provide not only high-throughput and low-latency wireless communications but also ultra-accurate and high-resolution wireless sensing capabilities. Such dual-functionality  is expected to significantly enhance spectrum efficiency, reduce hardware cost, and enable a wide range of emerging applications, including intelligent transportation, autonomous driving, extended reality, and smart manufacturing \cite{gonzelez2024integrated}.

These advantages have motivated extensive research efforts.
Existing ISAC studies have investigated the integration of sensing and communication from several complementary perspectives, including resource allocation and beamforming \cite{huahaocheng2023optimal,hua2023joint,zhao2024joint}, waveform design \cite{limimo2025,liuvortex2023,wang2025ofdm}, signal processing for delay and Doppler estimation \cite{hua2024integrated,wu2025low,yang2024sensing}, information-theoretic performance limits \cite{xiong2023onthefundamental, ren2024fundamental,wang2024unified}, network-level architecture and protocol design \cite{meng2025cooperative,jiang2025networklevel,meng2025network}, and learning-driven ISAC frameworks \cite{temiz2025deep,jiang2024ISACnet,lu2024deep}. These works have demonstrated the potential of jointly exploiting communication signals for data transmission and environmental sensing, enabling applications such as vehicular networks, autonomous driving, and smart transportation. Nevertheless, the majority of existing ISAC studies rely on wireless signals as the information source, making ISAC systems inherently sensitive to the wireless propagation environment. For instance,  in channels with severe multipath fading, the absence of a dominant line-of-sight (LoS) component may lead to drastic degradation in sensing accuracy.

An emerging  line of work leverages
 out-of-band information,  mostly  camera imagery,  to assist both communication and sensing. Environment information, such as  blockages,  can be easily obtained by cameras, thereby enhancing  ISAC performance in complex real-world environments \cite{charan2021vision6G,feng2024visionultra,xu2025computerscheudling, huang2024vision,xu2023computer,lin2024multicamera,wen2023vision,jiang2022computer,zhang2022vision}.   For example, recent studies \cite{charan2021vision6G,feng2024visionultra,xu2025computerscheudling} have demonstrated the potential of vision-assisted communication systems, where computer vision techniques are employed for blockage prediction. By proactively identifying potential obstacles, these schemes enable more reliable handoff management and efficient communication scheduling, thereby preventing link outages and improving the overall robustness of wireless networks. In addition, several recent works \cite{ huang2024vision,xu2023computer,lin2024multicamera,wen2023vision,jiang2022computer,zhang2022vision} have leveraged visual information from communication users in captured images, where the pixel-level coordinates of the users are detected and then utilized as input features. By employing deep learning techniques to learn the mapping between pixel coordinates and beam indices, these approaches effectively enable beam alignment and tracking, significantly reducing beam training overhead and improving the reliability of millimeter-wave and terahertz communication systems.  Furthermore, by simultaneously fusing multiple types of information, such as wireless signals, camera images, thermal infrared data, light detection and ranging (LiDAR) point clouds, and global positioning system (GPS)  information, ISAC systems can exploit complementary sensing modalities to overcome the limitations of single-source approaches \cite{yang2024semantic,zecchin2022lidar,chran2022vision,sagduyu2024jointsensing,charan2022multi,gao2025vision,zhang2025multimodal}.

However, existing single-modality ISAC solutions still  face two fundamental 
limitations  in multi-target scenarios:  (i) the ambiguity in associating 
delay and Doppler estimates with their corresponding targets; and (ii) the inability to distinguish multiple closely located targets due to limited system resources.   To the best of our knowledge, no prior work has resolved these limitations jointly.
Motivated by these challenges,  this paper develops a vision-assisted OFDM ISAC framework that leverages multi-modal fusion of wireless and visual data to enhance multi-target perception and parameter estimation. Specifically, the proposed system integrates deep joint source-channel coding (DeepJSCC)-based image transmission, YOLOv5-based object detection, and delay–Doppler map (DDM) feature extraction into a unified deep learning architecture for joint estimation of target location, transmission delay, and Doppler frequency.
The main contributions of this paper are summarized as follows:
\begin{itemize}
	\item  \textbf{Vision-assisted ISAC framework:} We propose a novel OFDM-based ISAC framework that incorporates visual sensing into the wireless sensing pipeline. By transmitting and reconstructing camera images via DeepJSCC, semantic visual information is extracted using a YOLOv5 detector to localize targets within the reconstructed images. Each detected target is represented by its associated visual features (e.g., bounding box and category) and fused with wireless features in a multi-modal network, thereby producing target-specific parameter estimates. This integration effectively resolves the ambiguity in multi-target association and mitigates the inability to distinguish multiple physically located target signals inherent in conventional ISAC methods.
	\item \textbf{Multi-modal feature embedding and fusion:} We design a deep neural network (DNN)  that embeds heterogeneous features, including bounding box coordinates, vehicle type, prior geometry information, and DDM features. These embeddings are then fused through convolutional neural networks (CNNs) and fully connected (FC) networks to produce accurate multi-parameter estimates.
\item \textbf{Novel loss function design:} To address the challenge of large output dimensions in delay–Doppler classification, we adopt a Kullback–Leibler (KL) loss against triangular soft labels, a variant of label distribution learning, to stabilize training of the high-dimensional delay/Doppler classifiers.
	\item \textbf{Realistic evaluation platform:} To validate the effectiveness of the proposed design, we build a Blender-based street-scene generator, producing tuples consisting of an image, channel state information (CSI), and ground-truth target state, for end-to-end evaluation.
 The results demonstrate that the proposed vision-assisted multi-modal learning framework not only achieves highly accurate parameter estimation but also overcomes the two fundamental limitations of conventional ISAC, i.e., ambiguity in multi-target association and the inability to distinguish multiple physically located target signals.
\end{itemize} 
The remainder of this paper is organized as follows. Section II presents the system model. Section III describes the signal processing techniques for multi-modal data fusion. Section IV introduces the proposed vision-assisted OFDM ISAC framework and details the corresponding neural network architecture. Section V provides numerical results to demonstrate the effectiveness of the proposed approach. Finally, Section VI concludes the paper.


 
\begin{figure}[!t]
	\centerline{\includegraphics[width=3in]{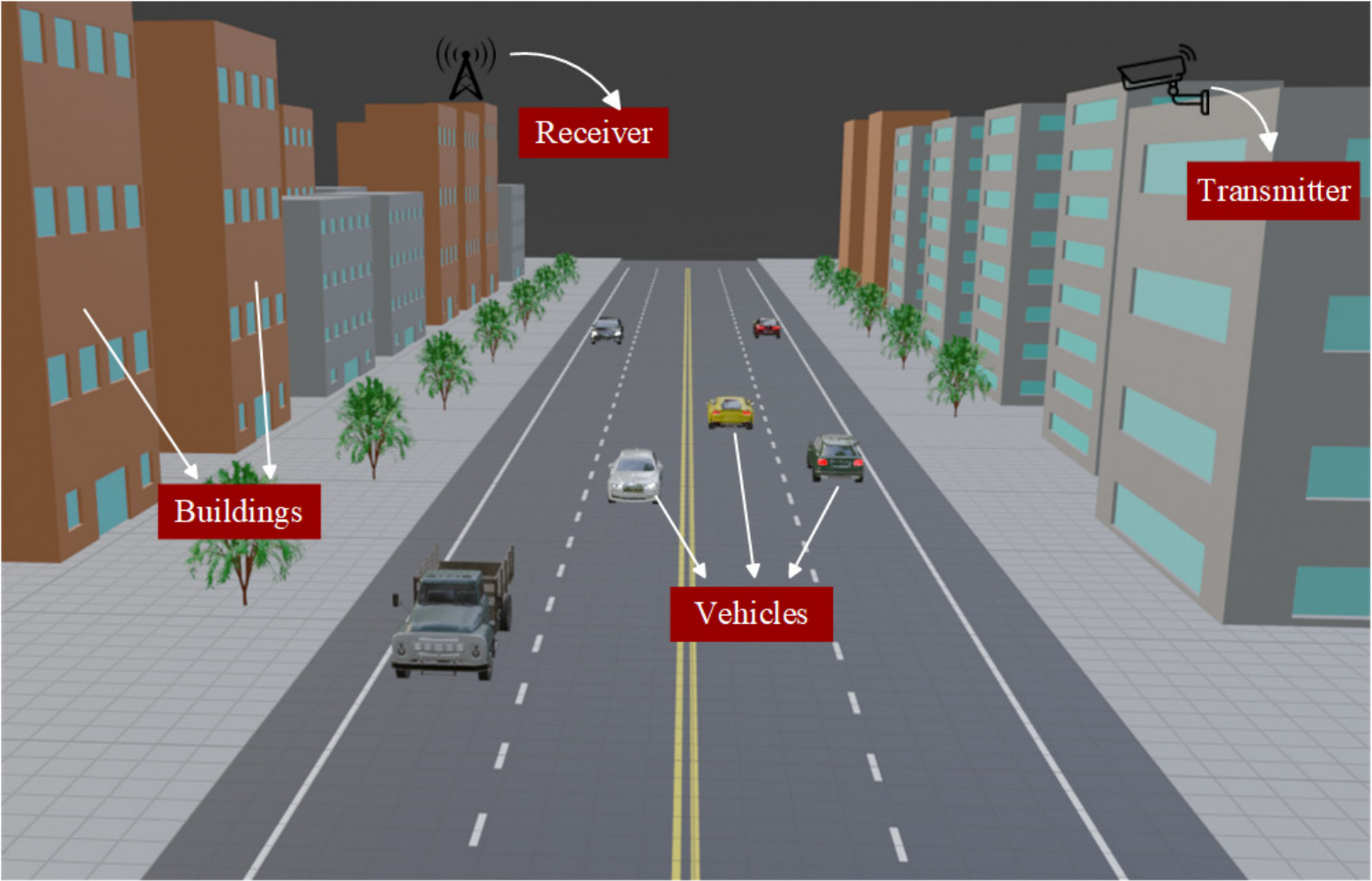}}
	\caption{Vision-assisted OFDM ISAC system model, where the transmitter employs an onboard camera to capture street-view images that are transmitted to the receiver via DeepJSCC, and the receiver reconstructs the images, detects targets, and fuses the visual information with wireless sensing features.} \label{systemmodel}
\end{figure}

\begin{figure*}[!t]
	\centerline{\includegraphics[width=5.5in]{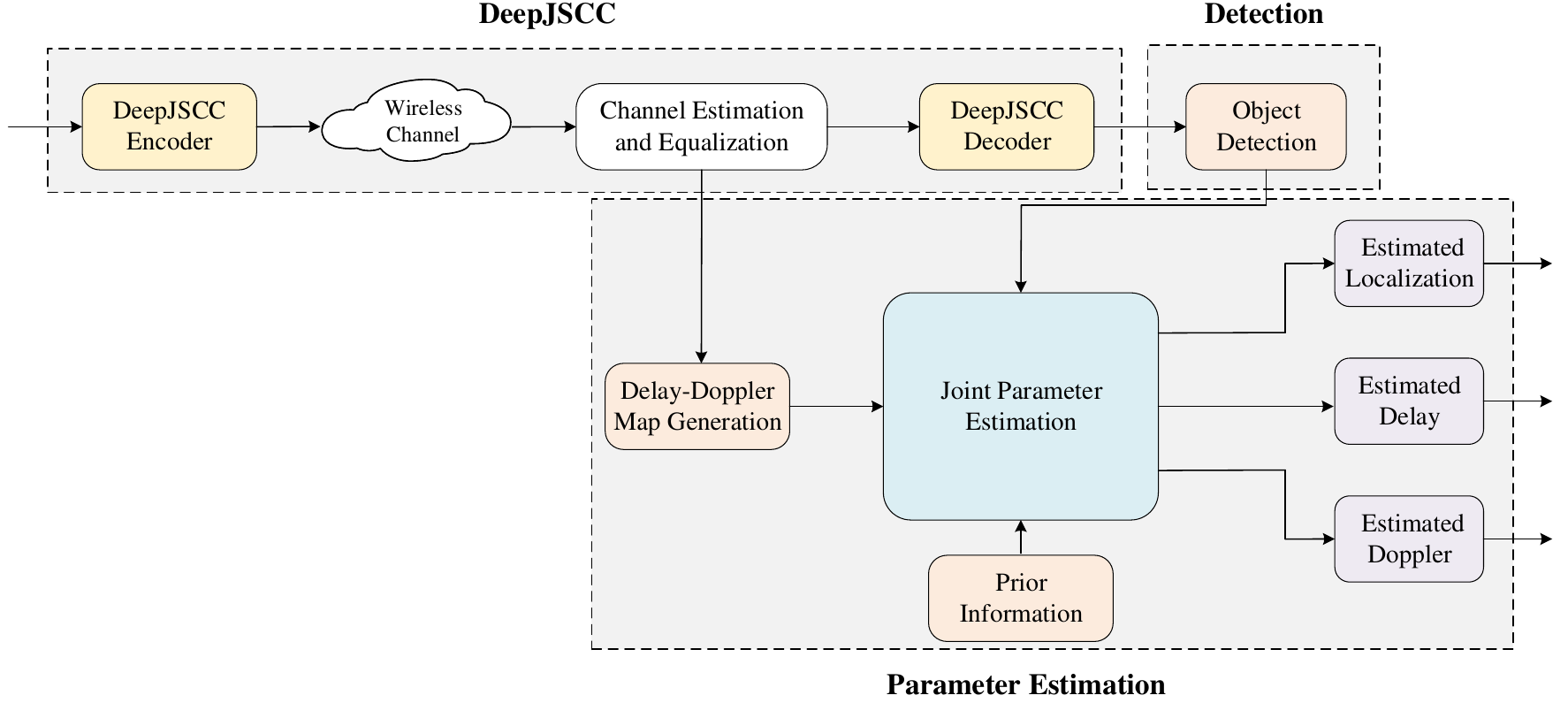}}
	\caption{The  diagram of implementing the proposed scheme.} \label{pipeline_overall_system}
\end{figure*}

\section{System Model}
We consider a vision-assisted OFDM ISAC system, as shown in Fig.~\ref{systemmodel}, where 
the transmitter is equipped with an onboard camera to capture street-view images of the surrounding environment. Each image is first processed by a DeepJSCC encoder, which extracts a compact feature vector representation. This feature vector is then transmitted over the wireless channel using OFDM signals instead of directly sending pixel values, thereby improving robustness against channel impairments. 
Each source image (after encoding) is carried within one OFDM packet, consisting of  $N_{\rm p}$  pilot symbols and $N_{\rm s}$ information symbols. Each  OFDM symbol
occupies  $N_{\rm c}$ subcarriers.   We
denote  the subcarrier spacing  by $\Delta f$ and   the OFDM symbol duration by $T_{\rm s}$, which satisfy $\Delta f = \frac{1}{T_{\rm s}}$. To mitigate inter-symbol interference, a cyclic prefix (CP) of duration $T_{\rm cp}$  is appended to each OFDM symbol, yielding a total symbol duration of  $T=T_{\rm s}+T_{\rm cp}$. For notational convenience, we define the pilot symbol index set, the subcarrier index set, and the target index set as  ${{\cal N}_{\rm{p}}} = \left\{ {0,1, \ldots ,{N_{\rm{p}}} - 1} \right\}$, ${{\cal N}_{\rm{c}}} = \left\{ {0,1, \ldots ,{N_{\rm{c}}} - 1} \right\}$, and ${{\cal K}} = \left\{ {1, \ldots ,K} \right\}$, respectively.

At the receiver side, three tasks are performed as illustrated in Fig.~\ref{pipeline_overall_system}.  First, the received feature vector is equalized and  fed into the DeepJSCC decoder. A YOLOv5 detector is then applied to the reconstructed street view to obtain semantic information of the targets, such as bounding boxes and vehicle categories. Second, the reflected OFDM signals from surrounding vehicles  are exploited to generate a DDM, which provides low-level physical features, including delay and Doppler shifts. 
Finally, the object-level visual features extracted from YOLOv5, prior knowledge (i.e., transmitter/receiver locations), and the DDM features are embedded and fused through a neural network. This multi-modal fusion enables joint estimation of each target’s three-dimensional  location, propagation delay, and Doppler-derived  velocity.
\section{Signal Processing for  Data Fusion}
In this section, we present the signal processing modules that extract wireless-domain sensing features for fusion with visual information. Specifically, we first introduce the \emph{channel estimation} procedure, which exploits OFDM pilot symbols to obtain the frequency-domain CSI. We then describe \emph{channel prediction and equalization}, in which the channel coefficients for data-bearing OFDM symbols are predicted from pilot-based estimates. The predicted channels are subsequently employed in minimum mean square error (MMSE) equalization to recover the transmitted symbols. Finally, we present the \emph{DDM generation}, which transforms the estimated channel responses into a structured two-dimensional representation that captures delay and Doppler shifts of surrounding targets. 
\subsection{Channel Estimation}
 Let ${{s}}_{n,p}^{{\rm{pilot}}} \in {{\mathbb C}^{1 \times 1}}$   denote the frequency-domain  pilot  symbol  on the $n$-th subcarrier of the $p$-th OFDM symbol. By applying the inverse discrete Fourier transform (IDFT) across all subcarriers, the frequency-domain pilot vector is transformed into a time-domain OFDM pilot waveform, which  can be expressed as \cite{li2025mimo}
 \begin{align}
{x_{\rm{p}}}(t) = \sum\limits_{p = 0}^{{N_{\rm{p}}} - 1} {\sum\limits_{n = 0}^{{N_{\rm{c}}} - 1} {s_{n,p}^{{\rm{pilot}}}} } {e^{j2\pi n\Delta f(t - {T_p})}}{\rm{rect}}\left( {\frac{{t - T_p}}{{{T}}}} \right),\label{pilotsequence}
 \end{align}
 where  ${T_p} \overset{\triangle}{=} pT$  and  ${\rm{rect}}\left(  \cdot  \right)$ is a rectangular pulse.
%
%
 
 The signal received by the receiver can be written as\footnote{We assume that static  clutter has been suppressed offline via background subtraction and focus on the residual target-induced channel.
} 
 \begin{align}
y(t) = \sum\limits_{k = 0}^K {{h_k}} {x_{\rm{p}}}(t - {\tau _k}) {e^{j2\pi {\nu _k}t}}  + z(t), \label{channeloutput}
 \end{align}
where $h_0$ denotes  the channel coefficient of the direct link  and  $h_k$ for $k \in {\cal K}$ represents the channel  coefficient associated with target $k$. 
The parameter $\tau_k$ corresponds to the propagation delay of the $k$-th target, while $\nu _k$ denotes the Doppler frequency shift induced by its relative radial velocity. The term  $z(t)$ represents  additive white Gaussian noise  with  zero mean and variance $\sigma^2$.

 In radar systems, millimeter-wave frequencies are typically employed to achieve high-resolution sensing. At such frequencies, non-LoS  propagation is easily blocked, and thus the channel can be accurately modeled as a LoS link. Specifically, $h_0$ and $h_k$ for $k \in {\cal K}$ can be   expressed as \cite{hua20243d}
 \begin{align}
 	{h_0} = \sqrt {\frac{{{\lambda ^2}}}{{16{\pi ^2}d_0^2}}} {\mkern 1mu} {\kern 1pt} {e^{ - j\frac{{2\pi }}{\lambda }{d_0}}},
 \end{align}
 and 
  \begin{align}
{h_k} = \sqrt {\frac{{{\lambda ^2}{\mkern 1mu} {\kern 1pt} {\rm{RC}}{{\rm{S}}_k}}}{{64{\pi ^3}{\mkern 1mu} {\kern 1pt} d_{{\rm{t}},k}^2{\mkern 1mu} {\kern 1pt} d_{{\rm{r}},k}^2}}} {\mkern 1mu} {\kern 1pt} {e^{ - j\frac{{2\pi }}{\lambda }\left( {{d_{{\rm{t}},k}} + {d_{{\rm{r}},k}}} \right)}},
 \end{align}
 where $\lambda $ denotes the carrier wavelength,  $d_0$ is the distance between the transmitter and receiver, ${{d_{{\rm{t}},k}}}$ and ${{d_{{\rm{r}},k}}}$  represent the  distances from the transmitter to target $k$ and from the   target $k$ to the receiver, respectively, and ${{\rm{RC}}{{\rm{S}}_k}}$ denotes the radar cross section of target $k$.

 Substituting \eqref{pilotsequence} into \eqref{channeloutput}, we have 
 \begin{align}
 {{{y}}_{\rm{p}}}(t) &= \sum\limits_{k = 0}^K {\sum\limits_{p = 0}^{{N_{\rm{p}}} - 1} {\sum\limits_{n = 0}^{{N_{\rm{c}}} - 1} {{{{h}}_k}{{s}}_{n,p}^{{\rm{pilot}}}} } }  \times  {e^{j2\pi n\Delta f(t - {T_p} - {\tau _k})}} \notag\\
 &  \times {e^{j2\pi {\nu _k}t}}{\rm{rect}}\left( {\frac{{t - T_p - {\tau _k}}}{T}} \right) + z(t). \label{received_pilot}
 \end{align}
The received signal in \eqref{received_pilot} is shifted by
${T_p}$  to align the useful part of the symbol at time zero. Then, the CP is removed, and a rectangular window 
${\mathop{\rm rect}\nolimits} \left( {\frac{t}{T_{\rm s}}} \right)$ is applied to extract the interval 
$t \in \left[ {0,T_{\rm s}} \right]$. The resulting  $p$-th OFDM symbol is given by
 
 \begin{align}
 {{{y}}_{{\rm{p}}}}(p,t) = {{{y}}_{\rm{p}}}\left( {t + {T_p}} \right){\mathop{\rm rect}\nolimits} \left( {\frac{t}{T_{\rm s}}} \right). \label{received_pilot: signleofdm}
 \end{align}
 After sampling the received signals  at the  rate
 $B = {N_{\rm{c}}}\Delta f$, the resulting discrete-time $q$-th sample of the $p$-th OFDM in   \eqref{received_pilot: signleofdm}
 can be written as
 \begin{align}
 {{{y}}_{\rm{p}}}\left[ {p,q} \right] &=  \sum\limits_{k = 0}^K {\sum\limits_{n = 0}^{{N_{\rm{c}}} - 1} {{{{h}}_k}{{s}}_{n,p}^{{\rm{pilot}}}} } {e^{j2\pi n\Delta f(\frac{q}{B} - {\tau _k})}}{e^{j2\pi {\nu _k}\left( {\frac{q}{B} + {T_p}} \right)}}  \notag\\
 & +{{z}}\left[ {p,q} \right]. \label{pilot_q}
 \end{align}
 To avoid inter-carrier interference, the OFDM subcarrier spacing ${\Delta f}$ should be  significantly larger than the maximum Doppler frequency shift  $\left\{ {{\nu _k}} \right\}_{k = 1}^K$, i.e., $\Delta f \gg \mathop {\max }\limits_k {\nu _k}$, so that ${e^{j\frac{{2\pi {\nu _k}q}}{B}}} = {e^{j\frac{{2\pi {\nu _k}q}}{{N\Delta f}}}} \approx 1$. Then, expression \eqref{pilot_q} can be approximated as 
 \begin{align}
 {{{y}}_{{\rm{p}}}}\left[p,q \right]&\approx 
 \sum\limits_{k = 0}^K {\sum\limits_{n = 0}^{{N_{\rm{c}}} - 1} {{{{h}}_k}{{s}}_{n,p}^{{\rm{pilot}}}} } {e^{j\frac{{2\pi nq}}{{{N_{\rm{c}}}}}}}{e^{ - j2\pi n\Delta f{\tau _k}}}{e^{j2\pi {\nu _k}{T_p}}} \notag\\
 &+ {{n}}\left[ p,q \right].  \label{pilot_q_approx}
 \end{align}
 
 To estimate the channel frequency response, an $N_{\rm c}$-point discrete Fourier transform (DFT) is applied to each OFDM symbol to convert the
 discrete-time signal in \eqref{pilot_q_approx} into the frequency domain. Let  ${{\bar y}_{\rm{p}}}\left[ {p,n} \right] \in {{\mathbb C}^{1 \times 1}}$ denote the $n$-th subcarrier of the $p$-th OFDM, we have 
 \begin{align}
 &{{{{\bar y}}}_{{\rm{p}}}}\left[ p,n \right] = {\rm{DFT}}\left( {{{{y}}_{{\rm{p}}}}\left[p, q \right]} \right) = \frac{1}{N_{\rm c}-1}\sum\limits_{q = 0}^{{N_{\rm{c}}-1}} {{{{y}}_{{\rm{p}}}}\left[p, q \right]{e^{ - j\frac{{2\pi nq}}{N_{\rm c}}}}} \notag\\
 &  = \sum\limits_{k = 0}^K {\sum\limits_{n' = 0}^{{N_{\rm{c}}} - 1} {{{{h}}_k}{{s}}_{n',p}^{{\rm{pilot}}}} } {e^{ - j2\pi n'\Delta f{\tau _k}}}{e^{j2\pi {\nu _k}{T_p}}}\notag\\
 &\quad \times {\mkern 1mu} \left( {\frac{1}{N_{\rm c}}\sum\limits_{q = 0}^{{N_{\rm{c}}-1}} {{e^{j\frac{{2\pi \left( {n' - n} \right)q}}{N_{\rm c}}}}} } \right) + {{\bar z}}\left[ p,n \right] \notag\\
 & = {{\bar h}}\left[ {p,n} \right]{{s}}_{n,p}^{{\rm{pilot}}} + {{\bar n}}\left[p, n \right], \label{receviedsignalafterDFT}
 \end{align}
 where ${{\bar h}}\left[ {p,n} \right] \overset{\triangle}{=} \sum\limits_{k = 0}^K {{{{h}}_k}{e^{ - j2\pi n\Delta f{\tau _k}}}{e^{j2\pi {\nu _k}{T_p}}}} $ and $\bar z \left[ {p,n} \right] \overset{\triangle}{=} \frac{1}{N_{\rm c}}\sum\limits_{q = 0}^{{N_{\rm{c}}-1}} {z\left[ {p,q} \right]{e^{ - j\frac{{2\pi nq}}{N_{\rm c}}}}} $. Here, ${\bar h}\left[ {p,n} \right] $ can be interpreted as the channel frequency response of the $n$-th subcarrier in the $p$-th OFDM symbol. 
 As a result, the channel $\bar h \left[ {p,n} \right]$, $p \in {\cal N}_{\rm p}, n \in {\cal N}_{\rm c}$, can be simply estimated as  
\begin{align}
\tilde h \left[ {p,n} \right] = \frac{{{{\bar y}_{\rm{p}}}\left[ {p,n} \right]}}{{s_{n,p}^{{\rm{pilot}}}}}, \label{estimatedchannel}
\end{align}
which is the least-squares (LS) estimator of ${{\bar h}}\left[ {p,n} \right]$.
\subsection{Channel Prediction and  Equalization}
Let ${{s_{n,p}}}\in {{\mathbb C}^{1 \times 1}}$  represent  the  data-bearing symbol  on the $n$-th subcarrier of the $p$-th OFDM symbol. The average transmit power, denoted by $P$,  satisfies $\frac{1}{{{N_{\rm{c}}}{N_s}}}\sum\limits_{n = 0}^{{N_{\rm{c}}} - 1} {\sum\limits_{p = {N_{\rm{p}}}}^{{N_{\rm{p}}} + {N_s} - 1} {{{\left| {{{\rm{s}}_{n,p}}} \right|}^2}} }  = P$. Similar to \eqref{receviedsignalafterDFT}, 
 the frequency-domain received signal can be expressed as 
\begin{align}
{{\bar y}}\left[ {p,n} \right] = {{\bar h}}\left[ {p,n} \right]{s_{n,p}} + {{\bar z}}\left[ {p,n} \right],  {\kern 1pt}  &p = {N_{\rm{p}}}, \ldots ,{N_{\rm{p}}} + {N_{\rm{s}}}-1,\notag\\
& n = 0, \ldots ,{N_{\rm{c}}-1}. \label{datatransmission}
\end{align}
Note that the transmitted data symbol ${{s_{n,p}}}$  cannot be directly obtained from the estimated channel in   \eqref{estimatedchannel}, because  $\tilde h \left[ {p,n} \right]$ is only available for pilot-bearing symbols  $p \in {\cal N}_{\rm p}$ and the channel is generally time-varying across OFDM symbols, i.e., $\tilde h \left[ {p,n} \right]$  differs with $p$. To enable equalization for $p \ge {N_{\rm{p}}}$, the channel coefficients must first be predicted for these indices. While more sophisticated predictors (i.e., Wiener or Kalman filter) can be used, we adopt linear extrapolation as a lightweight baseline that suffices for the Doppler regime considered here.  Specifically, let ${y_0}\left[ n \right]$ and ${y_1}\left[ n \right]$  denote the complex channel frequency response values at the $n$-th subcarrier at the last two pilot symbol indices, which are given by 
\begin{align}
{y_0}\left[ n \right] = \tilde h\left[ {{N_{\rm{p}}} - 2,n} \right] ~~{\rm and} ~~{y_1}\left[ n \right] = \tilde h\left[ {{N_{\rm{p}}} - 1,n} \right].
\end{align}
The per-subcarrier slope is then computed as 
\begin{align}
\xi \left[ n \right] = {y_1}\left[ n \right] - {y_0}\left[ n \right].
\end{align}
For a data-bearing OFDM symbol with index $p \ge {N_{\rm{p}}}$, the predicted channel is expressed as
\begin{align}
{{\tilde h}_{{\rm{pred}}}}\left[ {p,n} \right] = {y_1}\left[ n \right] + \left( {p - {N_{\rm{p}}} + 1} \right) \times \xi \left[ n \right].
\end{align}
The predicted channel coefficients ${{\tilde h}_{{\rm{pred}}}}\left[ {p,n} \right]$ are subsequently employed in the  MMSE equalization stage to obtain  $ \tilde s_{n,p}^{{\rm{MMSE}}}$  for the data-bearing symbols, which is given by 
 \begin{equation}
 \tilde s_{n,p}^{{\rm{MMSE}}} = \frac{{\tilde h_{{\rm{pred}}}^ * [p,n]\bar y[p,n]}}{{|{{\tilde h}_{{\rm{pred}}}}[p,n]{|^2} + {\sigma ^2}}}.
 \end{equation}

\subsection{Delay-Doppler Map Generation}
Although the estimated channel frequency response ${\tilde h [p,n]}$ in \eqref{estimatedchannel} is  available, the direct link carries no target information and may degrade the sensing performance. To address this, the direct link component is removed before target sensing, and the target-only channel frequency response is expressed as
\begin{align}
\hat h [p,n] = \tilde h[p,n] - {h _0}{e^{ - j2\pi n\Delta f{\tau _0}}}{e^{j2\pi {\nu _0}{T_p}}}, \label{channelwithouttargetinformation}
\end{align}
with  ${{\nu _0}}=0$.

After obtaining the estimate ${\hat h [p,n]}$ in \eqref{channelwithouttargetinformation}, a  DDM is constructed by applying a 2D DFT across both the time  and frequency  dimensions. 
The DDM, denoted by ${{\bm \Gamma} _{{\rm{DDM}}}} \in {{\mathbb C}^{{N_{\rm{p}}} \times {N_{\rm{c}}}}}$,   is obtained by performing  IDFT along the frequency  dimension, followed by a DFT along the slow-time (Doppler) dimension.  Mathematically, given the estimated channel frequency response $ {\hat h[p,n]}$, where $p \in {{\cal N}_{\rm{p}}}$  indexes the slow-time snapshots and $n \in {{\cal N}_{\rm{c}}}$ indexes the  subcarriers, the DDM is computed as
\begin{align}
{{\bm \Gamma} _{{\rm{DDM}}}}[i,j] &= \left| {{\rm{DF}}{{\rm{T}}_{p \to i}}\left( {{\rm{IDF}}{{\rm{T}}_{n \to j}}\left\{ {\hat h[p,n]} \right\}} \right)} \right|\notag\\
& =\left| {\sum\limits_{p = 0}^{{N_{\text{p}}} - 1} {\sum\limits_{n = 0}^{{N_{\text{c}}} - 1} {\hat h} } [p,n]{e^{j2\pi \left( {\frac{{jn}}{{{N_{\text{c}}}}} - \frac{{ip}}{{{N_{\text{p}}}}}} \right)}}} \right|,
\end{align}
 where $i$ and $j$   are the discrete delay and Doppler indices in the DDM,  and their resolutions are determined by $\Delta \tau  = \frac{1}{{{N_{\rm{c}}}\Delta f}}$ and $\Delta \nu  = \frac{1}{{{N_{\rm{p}}}\left( {{T_{\rm{s}}} + {T_{{\rm{cp}}}}} \right)}}$, respectively.
 The resulting magnitude spectrum 
${{\bm \Gamma} _{{\rm{DDM}}}}$ forms a 2D image (i.e., map), where each pixel reflects the signal energy received from a particular delay-Doppler bin. Peaks in this map indicate the presence of reflecting objects along with their corresponding propagation delays and Doppler shifts.

In conventional radar signal processing, classical approaches such as peak search algorithms are commonly employed to estimate delay–Doppler bins. These methods are effective in single- or isolated-target scenarios, where the dominant peaks in the DDM can be reliably associated with individual reflectors. However, in multi-target environments, a fundamental limitation arises: the DDM itself does not provide inherent information regarding the correspondence between a particular delay–Doppler pair and a specific physical target. Moreover, due to the finite system resolution, multiple targets with similar delays or Doppler shifts may produce closely spaced or overlapping signatures that cannot be distinguished by conventional energy-based techniques. As a result, whether the targets’ signatures are well separated, partially overlapping, or nearly coincident, traditional methods alone are unable to resolve the target association problem. This ambiguity is intrinsic to the DDM representation, stemming from the absence of explicit target identity information in the transformation process. These two limitations motivate our visually-augmented approach, which we present next.

\section{The Proposed Parameter Estimation Framework}
In this section, we present a vision-assisted parameter estimation framework that integrates visual and wireless sensing to enhance ISAC performance. First, the transmitter encodes street-view images into feature vectors and transmits  over the wireless channel using DeepJSCC. Second, YOLOv5 is applied to the reconstructed images at the receiver for target detection and extraction, yielding semantic features such as bounding boxes and categories. Finally, a multi-modal data fusion architecture combines visual features with the DDM and prior information through a DNN, enabling accurate estimation of target localization, delay, and velocity.

\begin{figure*}[!t]
	\centerline{\includegraphics[width=7in]{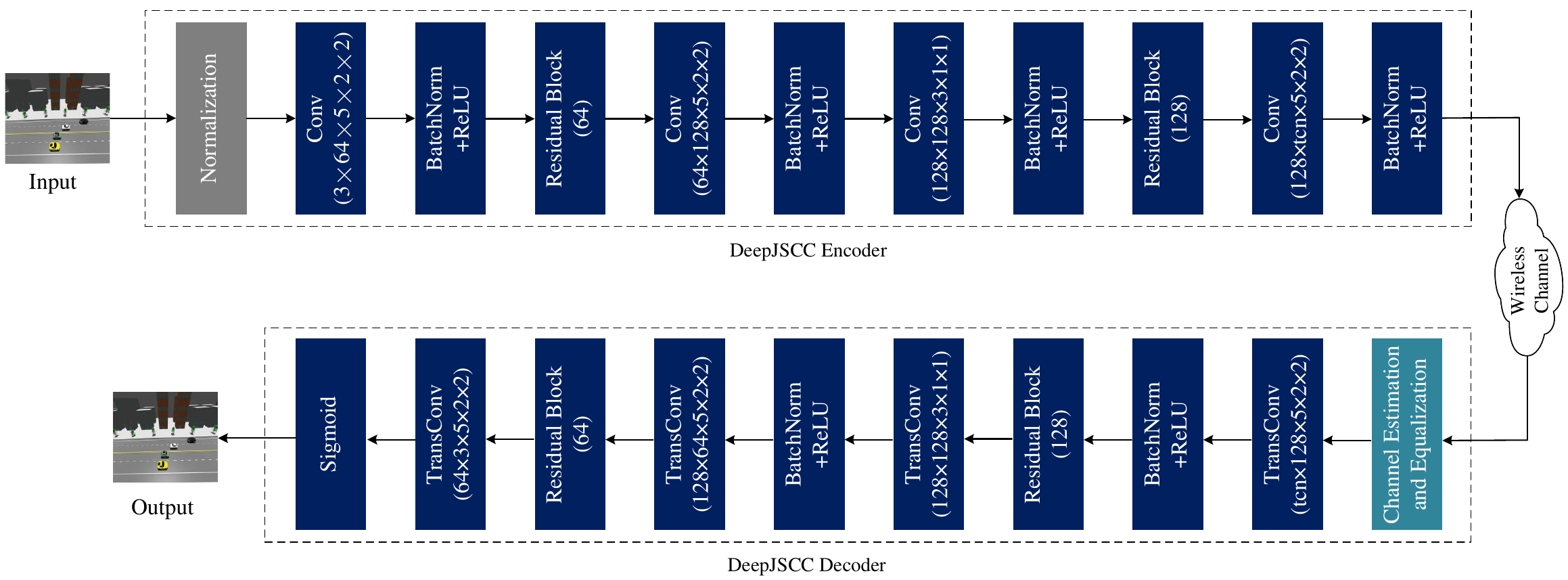}}
	\caption{The  pipeline of implementing DeepJSCC for high-resolution image transmission.} \label{pipeline_DeepJSCC}
\end{figure*}
\begin{figure*}[!t]
	\centerline{\includegraphics[width=7in]{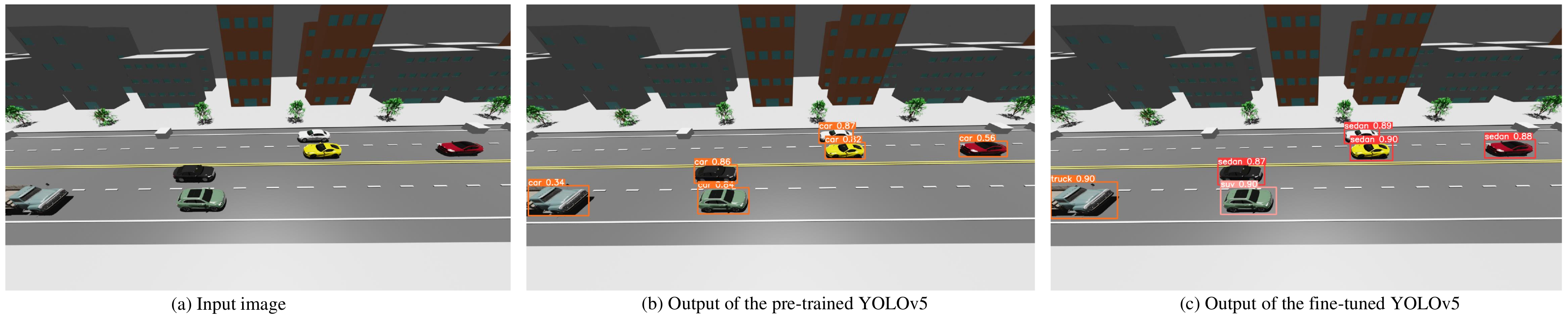}}
	\caption{A visualization of the output from YOLOv5 object detector.} \label{YOLOv5_detector}
\end{figure*}
\subsection{DeepJSCC for Image Transmission}
We employ DeepJSCC for image transmission, as it has been shown to outperform  conventional  separation-based image transmission,  particularly at a low signal-to-noise ratio (SNR) and  at a low channel bandwidth ratio \cite{Bourtsoulatze2019deep}.
However,  the CNN-based DeepJSCC framework in \cite{Bourtsoulatze2019deep} struggles to extract sufficient high-level features from high-resolution images. To address this limitation, we redesign the network architecture as illustrated in Fig.~\ref{pipeline_DeepJSCC}.  Specifically, the encoder incorporates multiple convolutional (Conv) layers, batch normalization (BatchNorm) layers, rectified linear unit (ReLU) activations, and residual blocks to enhance feature extraction and stabilize training. A Conv layer with parameters ${C_1} \times {C_2} \times K \times S \times P$ denotes a convolution with input channel dimension ${C_1}$,  output channel dimension ${C_2}$, kernel size $K \times K$, stride $S$, and padding $P$. Here, the residual blocks are introduced to facilitate gradient flow in deeper networks, thereby improving representation learning for complex image features.  The decoder mirrors the encoder structure but replaces Conv layers with transposed convolutional (TransConv) layers, which progressively upsample and reconstruct the image. The parameters ${C_1} \times {C_2} \times K \times S \times P$ in the TransConv layer are the same as in the encoder. 
\subsection{Target Detection and Extraction}
The receiver extracts vehicle-related information from the reconstructed image. Since a single image may contain multiple vehicles, object detection is required. The detection process serves two main purposes: \textit{1) Category:} The detection model must identify and classify the vehicles, as different vehicle types exhibit distinct RCS values that significantly affect the characteristics of the DDM; \textit{2) Location:} The detection model must provide the bounding box coordinates of vehicles, which implicitly reflect their true physical positions.

We adopt the advanced YOLOv5 detector for object detection and classification \cite{yolov5}. The YOLOv5 architecture comprises multiple convolutional layers, including backbone, neck, and head modules. The backbone extracts features from the input image,  the neck fuses shallow spatial and deep semantic features, and the head subsequently utilizes these features to predict the category and location of objects. For our specific application scenario, the target set is restricted to three vehicle classes: sedan, SUV, and truck.  Instead of training from scratch, we fine-tune the YOLOv5  using a small number of specific street samples to adapt the detector to our visual environment and target categories. A visualization of the output from pre-trained YOLOv5 and fine-tuned YOLOv5 is shown in Fig.~\ref{YOLOv5_detector}, where  Fig.~\ref{YOLOv5_detector}(a) is the original image as the input of YOLOv5 detector, while Fig.~\ref{YOLOv5_detector}(b) and Fig.~\ref{YOLOv5_detector}(c) denote the output results of pre-trained YOLOv5 and fine-tuned YOLOv5, respectively. The  YOLOv5 detector is pre-trained on the COCO dataset \cite{lin2014microsoft}, which contains 143,000 images with 80 object classes. We further fine-tune it using 500 specific street images. It can be observed that the pre-trained YOLOv5 in Fig.~\ref{YOLOv5_detector}(b)  fails to classify the vehicles, in contrast, the fine-tuned YOLOv5 can accurately detect and classify vehicles as shown in Fig.~\ref{YOLOv5_detector}(c). 
Note that each bounding box localized by YOLOv5 is parameterized by four key attributes: center coordinates, width, and height, which are normalized to the range $\left[ {0,1} \right]$ relative to the input image dimensions.  

%

\begin{figure*}[!t]
	\centerline{\includegraphics[width=5in]{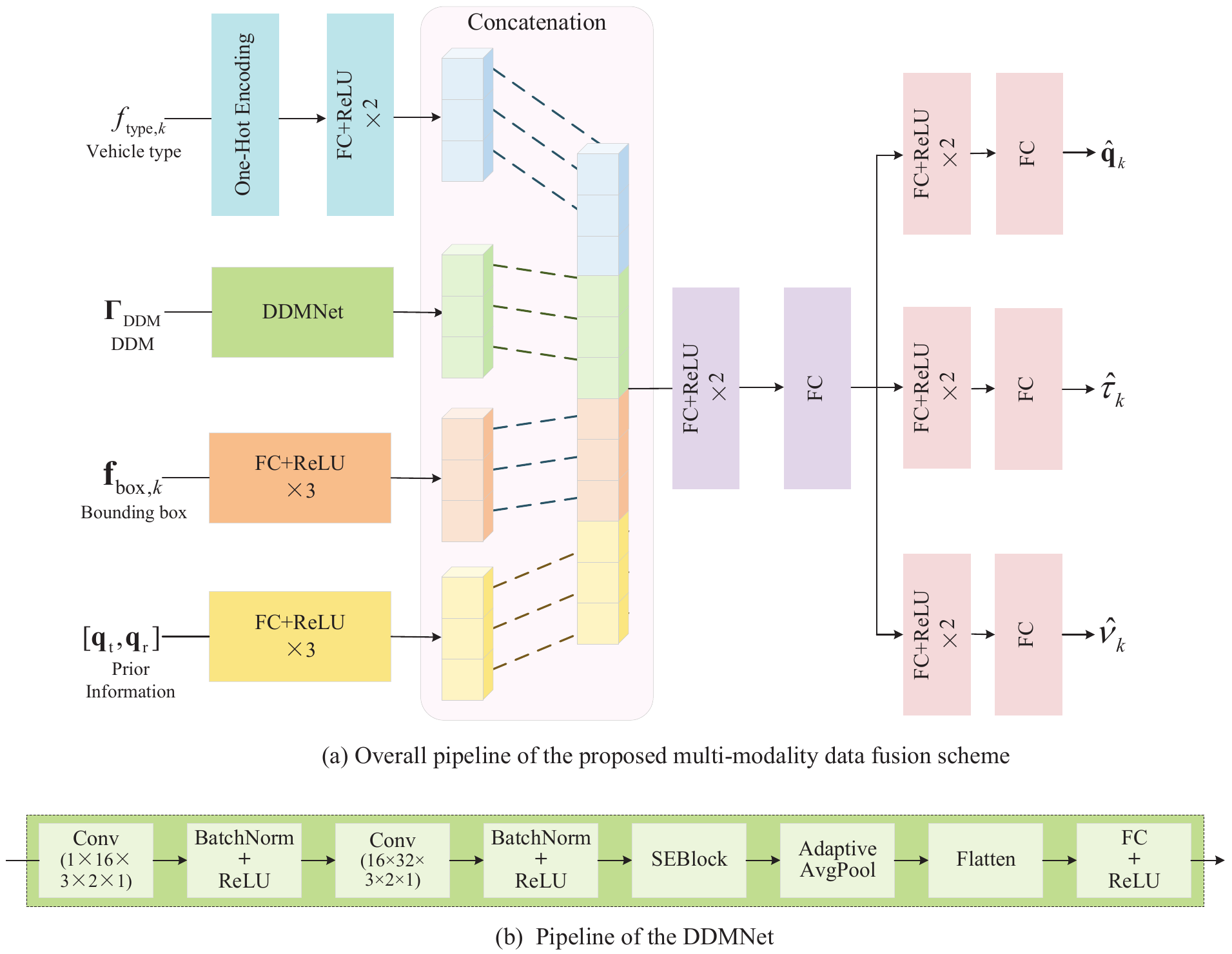}}
	\caption{Overall pipeline of the proposed multi-modality data fusion scheme.} \label{Overall_architecture_pipeline}
\end{figure*}

\subsection{Multi-Modal Data Fusion Architecture}
The overall pipeline of the proposed multi-modal data fusion scheme is shown in Fig.~\ref{Overall_architecture_pipeline}, where the object detection results, the DDM, and  prior information are jointly fed into  a neural network-based parameter estimation module. This module generates three key outputs: estimated target location, estimated delay, and estimated Doppler. 

\subsubsection{Vehicle Type Embedding}
To effectively incorporate categorical vehicle-type information into the joint parameter estimation framework, we design a lightweight embedding network, as illustrated in Fig.~\ref{Overall_architecture_pipeline}(a). The input vehicle type feature ${f_{{\rm{type}},k}} \in \left\{ {0,1,2} \right\}$
is first transformed into a one-hot vector, where each dimension corresponds to a specific vehicle type, i.e., sedan, SUV, and truck.
The one-hot encoded vector is then passed through two FC layers with ReLU activation.  Different vehicle types exhibit distinct  RCS characteristics, which directly affect their interaction with the wireless channel. By explicitly incorporating vehicle-type information, the embedding facilitates the network to better learn the channel environment, thereby improving its ability to infer the underlying vehicle state information, such as delay and Doppler.
Mathematically,  the embedding process is expressed as
\begin{align}
{{\bf e}_{{\rm{type,}}k}} = {g_{{\bm \theta}_1 }}\left( {{f_{{\rm{type}},k}}} \right),
\end{align}
where ${g_{{\bm \theta}_1} }\left(  \cdot  \right)$ denotes the vehicle type embedding function parameterized by DNN parameters ${\bm \theta}_1$.

\subsubsection{DDM Feature Extraction} 
The DDM provides a two-dimensional representation of multipath reflections, encapsulating propagation delays and Doppler shifts. To exploit this structured information, we design a CNN-based feature extractor, as illustrated in Fig.~\ref{Overall_architecture_pipeline}(b). The input DDM matrix is processed through successive convolutional layers with batch normalization and ReLU activations to capture local temporal spectral correlations, followed by a squeeze-and-excitation (SE) block that adaptively reweights feature channels. An adaptive average pooling layer and an FC layer are then applied to obtain a compact embedding vector.
Mathematically,  the feature function parameterized by DNN parameters ${\bm \theta}_2$ is expressed as
\begin{align}
	{\bf{e}}_{\rm DDM} = {g_{{{\bm \theta} _2}}}\left( {{{\bm \Gamma} _{{\rm{DDM}}}}} \right).
\end{align}

\subsubsection{Bounding Box Feature Embedding}
As shown in Fig.~\ref{Overall_architecture_pipeline}(a), the bounding box feature vector ${{\bf{f}}_{{\rm{box}},k}} = \left[ {x\text{-}{\rm{center}},y\text{-}{\rm{ center}},{\rm{width,height}}} \right] \in {{\mathbb R}^{4 \times 1}}$ is processed through three FC layers, each followed by a ReLU activation. 
The bounding box embedding plays an important role in bridging the image domain with the physical environment. Specifically, the pixel-level bounding box information obtained from object detection is mapped through the embedding network into meaningful physical representations. This allows the network to exploit image-domain cues for accurate vehicle localization in the real world.
Mathematically,  we can express this embedding process as 
\begin{align}
{{\bf e}_{{\rm{box,}}k}} = {g_{{\bm \theta}_3 }}\left( {{{\bf f}_{{\rm{box}},k}}} \right),
\end{align}
where ${g_{{\bm \theta}_3} }\left(  \cdot  \right)$ denotes the bounding box embedding function parameterized by DNN parameters ${\bm \theta}_3$.

\subsubsection{Prior Information Embedding}
In addition to visual and signal-domain features, prior knowledge  of the physical configuration provides valuable cues for parameter estimation. In our framework, the prior information corresponds to the physical locations of the transmitter and receiver,  serving as auxiliary inputs to enhance transmission delay estimation accuracy. As illustrated in Fig.~\ref{Overall_architecture_pipeline}(a), the prior information vector is also passed through a sequence of three  FC layers, each followed by a  ReLU activation. 
Since the delay and velocity are functions of the physical locations of the transmitter and receiver,  the framework effectively reduces estimation ambiguity by incorporating this prior knowledge.
Mathematically,  this feature extraction process is expressed as
\begin{align}
{{\bf{e}}_{{\rm{position}}}} = {g_{{{\bm \theta} _4}}}\left( {{\rm{concat}}[{{\bf{q}}_{\rm t}},{{\bf{q}}_{\rm r}}]} \right),
\end{align}
where ${{\bf{q}}_{\rm t}} \in {{\mathbb R}^{3 \times 1}}$ and ${{\bf{q}}_{\rm r}} \in {{\mathbb R}^{3 \times 1}}$ denote the locations of transmitter and receiver, respectively, and ${g_{{\bm \theta}_4} }\left(  \cdot  \right)$ denotes the transmitter and receiver location embedding function parameterized by DNN parameters ${\bm \theta}_4$.

\subsubsection{Feature Embedding Fusion} 
After obtaining modality-specific embeddings from the vehicle type, bounding box, DDM, and prior information, we concatenate these latent vectors into a single joint representation,  as illustrated in Fig.~\ref{Overall_architecture_pipeline}(a).  The fused embedding is subsequently passed through a neural network comprising multiple FC layers with ReLU activations. This network is designed to exploit cross-modal correlations among the different embeddings and to learn higher-order interactions that are critical for accurate parameter estimation. Mathematically, this process can be expressed as
\begin{align}
&{{{{\bf{\hat q}}}_k}},\;{\kern 1pt} {{\hat \tau }_k},\;{\kern 1pt} {{\hat \nu}_k} = \notag\\
&\qquad {g_{{{\bm \theta} _5}}}\left( {{\rm{concat}}[{{\bf{e}}_{{\rm{type}},k}},\;{\kern 1pt} {{\bf{e}}_{{\rm{box}},k}},\;{\kern 1pt} {{\bf{e}}_{{\rm{DDM}}}},\;{\kern 1pt} {{\bf{e}}_{{\rm{position}}}}]} \right),
\end{align}
where ${g_{{\bm \theta}_5} }\left(  \cdot  \right)$ represents the FC-based fusion network parameterized by ${{{\bm \theta} _5}}$. The outputs 
${{\bf {\hat {q}}}_k}\in {\mathbb R}^{3 \times 1}$, $ {{\hat \tau }_k}\in {\mathbb R}^{1 \times 1}$, and $ {{\hat \nu }_k}\in {\mathbb R}^{1 \times 1}$ represent the estimated vehicle $k$'s location, delay, and  Doppler, respectively.


\subsection{Loss Function and Overall Algorithm}
It is worth noting that the proposed framework adopts a modular training strategy, where the DeepJSCC, the YOLOv5 detector, and the multi-modal fusion network are trained independently. Specifically, the DeepJSCC module adopts the same mean square error (MSE)-based training loss function as in \cite{Bourtsoulatze2019deep}.
The output of the fine-tuned YOLOv5 detector is directly utilized as the ground-truth labels for supervising the multi-modal fusion network. 

The fusion network produces three outputs, namely, vehicle location, delay, and Doppler. The estimated vehicle location is represented as a vector with continuous-valued entries, whereas the delay and Doppler are represented as discrete integer values. Accordingly, different loss functions are employed for training. Specifically, we adopt the MSE-based loss function for estimating location, which is given by 
\begin{align}
{{\cal L}_{{\rm{location}},k}} = {\left\| {{{\bf{q}}_{{\rm{gt,}}k}} - {{{\bf{\hat q}}}_k}} \right\|^2},
\end{align}
where ${{{\bf{q}}_{{\rm{gt,}}k}}}$ is the ground-truth location of  vehicle $k$.

The outputs corresponding to delay and Doppler are discretized into hundreds of bins.  This formulation naturally converts delay and Doppler estimation into classification problems, where each bin represents a discrete delay or Doppler hypothesis. Thus,  the conventional cross-entropy loss with one-hot ground-truth labels can be adopted. However,  the cross-entropy loss often leads to unstable training and poor generalization when the output dimensions are very large. 

To overcome this limitation, we adopt the  KL divergence loss with soft target distributions.  Instead of a one-hot label, we construct a top-$L$ soft label that distributes probability mass over the ground-truth bin and its closest neighbors with a linearly decaying weight.  Specifically, the loss functions for delay and Doppler are defined as
 \begin{equation}
 	\mathcal{L}^{\tau}_{k} 
 	= D_{\mathrm{KL}}\!\big(\mathbf{q}^{(\tau)}_{k}\,\|\,\mathbf{p}^{(\tau)}_{k}\big) 
 	= \sum_{n=0}^{N_{\tau}-1} q^{(\tau)}_{k,n} \,
 	\log \frac{q^{(\tau)}_{k,n}}{p^{(\tau)}_{k,n}},
 \end{equation}
and 
 \begin{equation}
 	\mathcal{L}^{\nu}_{k} 
 	= D_{\mathrm{KL}}\!\big(\mathbf{q}^{(\nu)}_{k}\,\|\,\mathbf{p}^{(\nu)}_{k}\big) 
 	= \sum_{m=0}^{N_{\nu}-1} q^{(\nu)}_{k,m} \,
 	\log \frac{q^{(\nu)}_{k,m}}{p^{(\nu)}_{k,m}},
 \end{equation}
where $\mathbf{p}^{(\tau)}_{k}$ and $\mathbf{p}^{(\nu)}_{k}$ denote the predicted probability distributions obtained from the log-softmax outputs, and $\mathbf{q}^{(\tau)}_{k}$ and $\mathbf{q}^{(\nu)}_{k}$ are the  soft labels, which are defined as 
\begin{equation}
q_{k,n}^{(\tau)} = \frac{w_{|n - t^{(\tau)}_{k}|}}{\sum\limits_{j} w_{|j - t^{(\tau)}_{k}|}}, 
\quad 
q_{k,m}^{(\nu)} = \frac{w_{|m - t^{(\nu)}_{k}|}}{\sum\limits_{j} w_{|j - t^{(\nu)}_{k}|}},
\end{equation}
with 
\begin{equation}
w_{i} = 
\begin{cases}
1 - \dfrac{i}{L}, & i < L, \\[2mm]
0, & i \geq L.
\end{cases}
\end{equation}
The total training loss is defined as the weighted sum of the three component losses, expressed as
\begin{align}
\!\!{{\cal L}_{{\rm{total}},k}} = {w_1}{\mathbb E}\left\{ {{{\cal L}_{{\rm{location,}}k}}} \right\} + {w_2}{\mathbb E}\left\{ {{\cal L}_k^\tau } \right\} + {w_3}{\mathbb E}\left\{ {{\cal L}_k^\nu } \right\},
\end{align}
where the expectation is taken over all samples, and  $w_i > 0, i\in \{1,2,3\}$, are the weighting factors. 

\section{Numerical Results}

\subsection{Simulation Setup}
\subsubsection{Communication and Sensing Parameter Setup}
The simulation environment is configured with the transmitter and receiver placed at fixed positions. The coordinates of the transmitter are set to $(4.96,\,107,\,15)\,\text{m}$, while the receiver is located at $(43,\,0,\,20.5)\,\text{m}$. The RCS values of the considered vehicle types are specified as 10~dBsm for sedans, 15~dBsm for SUVs, and 25~dBsm for trucks.  The system operates at a carrier frequency of 28~GHz with a subcarrier spacing of 120~kHz.  The noise power is set to $\sigma^2 = -80~\text{dBm}$. The SNR is defined as \(\text{SNR}=\frac{P}{{{\sigma ^2}}}\). In addition, we set $w_1=w_2=w_3=1$.

\subsubsection{Image and Label Generation}
A synthetic dataset is constructed using the Blender software to emulate realistic street-view scenarios. In total, 2500 images are rendered at a resolution of 
$1920\times1080$ pixels. The road has a fixed width of 20 m and a length 120 m. The scene consists of two traffic directions with four lanes in total. The transmitter is mounted on the wall of a roadside building, while the receiver is placed on the rooftop of a building located on the opposite side of the street. Vehicle velocities are randomly sampled from the interval 
$\left[ {5,30} \right]$ m/s. Each rendered image is paired with corresponding object detection labels, providing bounding box and category information. The dataset is divided into two parts: 80\%  for training and    20\%  for testing.

\subsection{Evaluation Metrics}
Two evaluation metrics, namely Top-$K$ accuracy and root mean square error (RMSE),  are adopted.
\subsubsection{Top-$K$ accuracy} 
The Top-$K$ accuracy is a widely used performance metric for multi-class classification. It is defined as the percentage of test samples whose ground truth
bin index lies within the $K$ most likely estimated bins. Mathematically, 
 let $N$ denote the total number of samples, $y_n \in \{1,2,\dots,C\}$ the 
ground truth class label of the $n$-th sample, and 
$\mathbf{p}_n \in \mathbb{R}^{C \times 1}$ the predicted probability distribution over $C$ classes. 
Denote by $\text{Top}_{K}(\mathbf{p}_n)$ the set of class indices corresponding to the 
$K$ largest entries of $\mathbf{p}_n$. The top-$K$ accuracy is then defined as
\begin{equation}
	\text{Acc}_{\text{Top-}K} 
	= \frac{1}{N}\sum_{n=1}^{N} 
	\mathbf{1}\left\{ y_n \in \text{Top}_{K}(\mathbf{p}_k) \right\},
\end{equation}
where $\mathbf{1}\{\cdot\}$ is the indicator function that returns $1$ if the condition 
is satisfied and $0$ otherwise. When $K=1$, this metric reduces to the conventional 
classification accuracy.

Let $N_{{\tau}_i}$ and $N_{\hat {\tau}_i}$   denote the ground-truth and estimated delay bin indices of the $i$-th target, respectively, and $N_{{\nu}_i}$ and $N_{\hat {\nu}_i}$ denote the corresponding Doppler bin indices. The RMSE metric is computed based on  hard prediction obtained from the maximum a posteriori bin index, i.e.,
\begin{equation}
N_{\hat {\tau}_i} = \arg\max_j \; p_{\tau,i}(j), 
\qquad 
N_{\hat {\nu}_i} = \arg\max_j \; p_{\nu,i}(j),
\end{equation}
and then mapped into physical units using the corresponding delay and Doppler resolutions.  
\subsubsection{RMSE of Delay}
The delay RMSE is computed as the difference between the predicted and ground-truth delay bin indices, scaled by the delay resolution $\Delta\tau = \tfrac{1}{N_c \cdot \Delta f}$. Formally,
\begin{equation}
\text{RMSE}_{\text{delay}} = \Delta\tau \cdot \sqrt{\frac{1}{N}\sum_{i=1}^{N} \left( N_{{\tau}_i} - N_{\hat {\tau}_i} \right)^2 } .
\end{equation}

\subsubsection{RMSE of Doppler}
Similarly, the Doppler RMSE is calculated by scaling the bin index error with the Doppler resolution $\Delta \nu  = \frac{1}{{{N_{\rm{p}}}\left( {{T_{\rm{s}}} + {T_{{\rm{cp}}}}} \right)}}$. Thus, 
\begin{equation}
\text{RMSE}_{\text{doppler}} = \Delta\nu \cdot \sqrt{\frac{1}{N}\sum_{i=1}^{N} \left( N_{\hat {\nu}_i} - N_{{\nu}_i} \right)^2 } .
\end{equation}

\subsubsection{RMSE of Location}
The RMSE of the target location is computed as  
\begin{equation}
\text{RMSE}_{\text{location}} = \sqrt {\frac{1}{N}\sum\limits_{n = 1}^N {\left\| {{{\bf{q}}_{{\rm{gt}},n}} - {{{\bf{\hat q}}}_n}} \right\|^2} } .
\end{equation}

\subsubsection{RMSE of Velocity}
Based on the estimated Doppler frequency shift, the target velocity is further derived using bistatic geometry, as expressed in
\begin{align}
\nu  = \frac{{{{\bf{v}}^T}\left( {{{\bf{q}}_{{\rm{gt}},n}} - {{\bf{q}}_{\rm{t}}}} \right)}}{{\lambda \left\| {{{\bf{q}}_{{\rm{gt}},n}} - {{\bf{q}}_{\rm{t}}}} \right\|}} + \frac{{{{\bf{v}}^T}\left( {{{\bf{q}}_{{\rm{gt}},n}} - {{\bf{q}}_{\rm{r}}}} \right)}}{{\lambda \left\| {{{\bf{q}}_{{\rm{gt}},n}} - {{\bf{q}}_{\rm{r}}}} \right\|}}, \label{dopper_convert_velocity}
\end{align} 
where ${\bf v} \in {\mathbb R}^{3\times 1}$ denotes the vehicle velocity. Note that since the vehicles are constrained to move along the $y$-axis in either the positive or negative direction, the velocity estimation can be directly derived from \eqref{dopper_convert_velocity}. 
Accordingly, the velocity RMSE is given by
\begin{equation}
\text{RMSE}_{\text{velocity}} = \sqrt{\frac{1}{N}\sum_{i=1}^{N} \left\| \mathbf{v}_i - \hat{\mathbf{v}}_i \right\|^2 } ,
\end{equation}
where $\mathbf{v}_i$ and $\hat{\mathbf{v}}_i$ denote the ground-truth and estimated velocity vectors of the $i$-th target, respectively.

\begin{figure}[!t]
	\centering
	{\includegraphics[width=0.9\columnwidth]{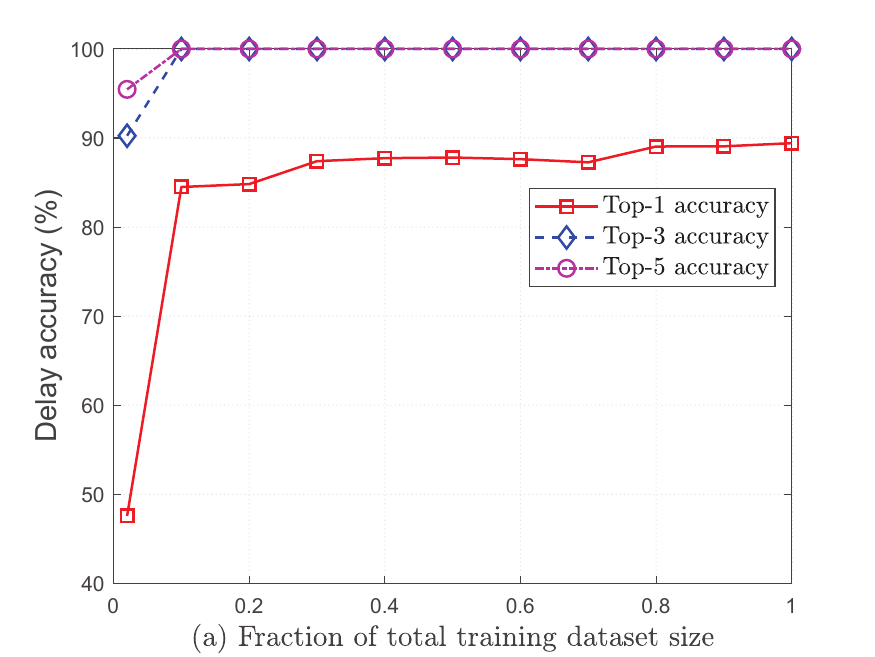}}\\
	{\includegraphics[width=0.9\columnwidth]{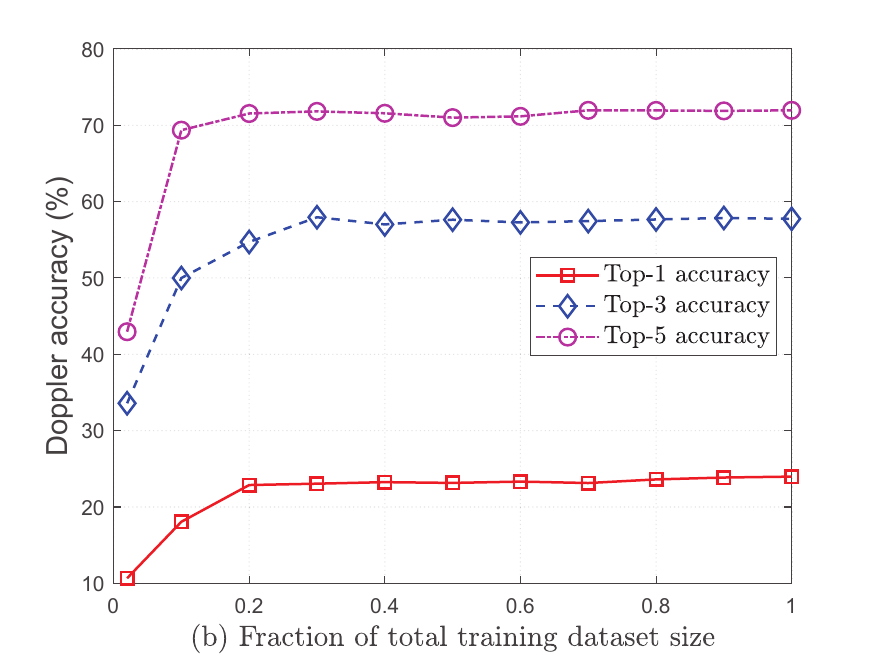}}
	\caption{Delay and Doppler accuracy estimation versus training dataset size.}
	\label{fig:delay_doppler_versus_dataset}
\end{figure}

\subsection{Performance Evaluation} \label{subsec:evaluation}

Fig.~\ref{fig:delay_doppler_versus_dataset} illustrates the impact of the training dataset size on the accuracy of delay and Doppler estimation for 
$N_{\rm c}=256$, $N_{\rm p}=512$, and $\text{SNR} = 20~\text{dB}$. As shown in Fig.~\ref{fig:delay_doppler_versus_dataset}(a) and Fig.~\ref{fig:delay_doppler_versus_dataset}(b), both delay and Doppler accuracies increase with the growth of the training dataset size. When the fraction of training data exceeds 30\% of the dataset, the performance remains nearly unchanged, indicating that the proposed scheme can achieve satisfactory results even with a limited amount of training data. 
 It is also observed that the overall Doppler accuracy is lower than that of the delay accuracy. This performance gap arises because the Doppler estimation network has an output dimension of 512, which is twice that of the delay estimation network, thereby making Doppler estimation inherently more challenging.

\begin{figure*}[!t]  
	\centering
	{
		\hspace{-1.2cm}	\includegraphics[width=0.38\textwidth]{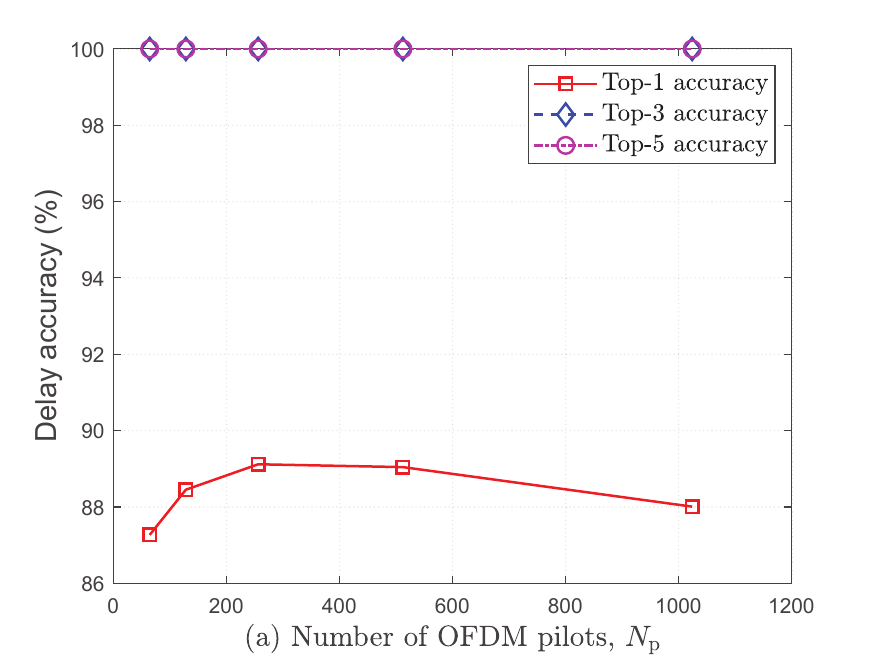}\hspace{-1cm}
	}
	{
		\includegraphics[width=0.38\textwidth]{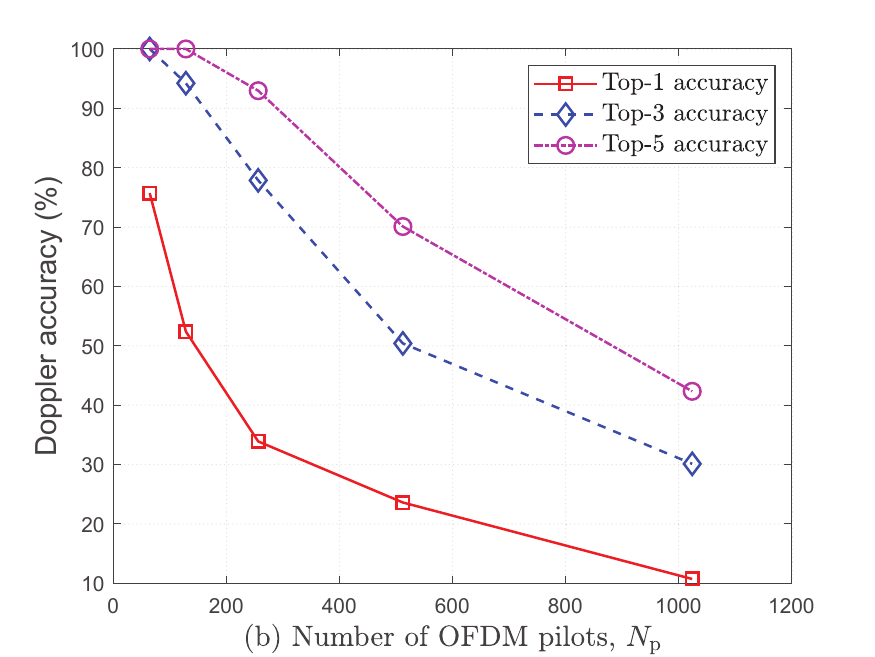}\hspace{-1cm}
	}
	{
		\includegraphics[width=0.38\textwidth]{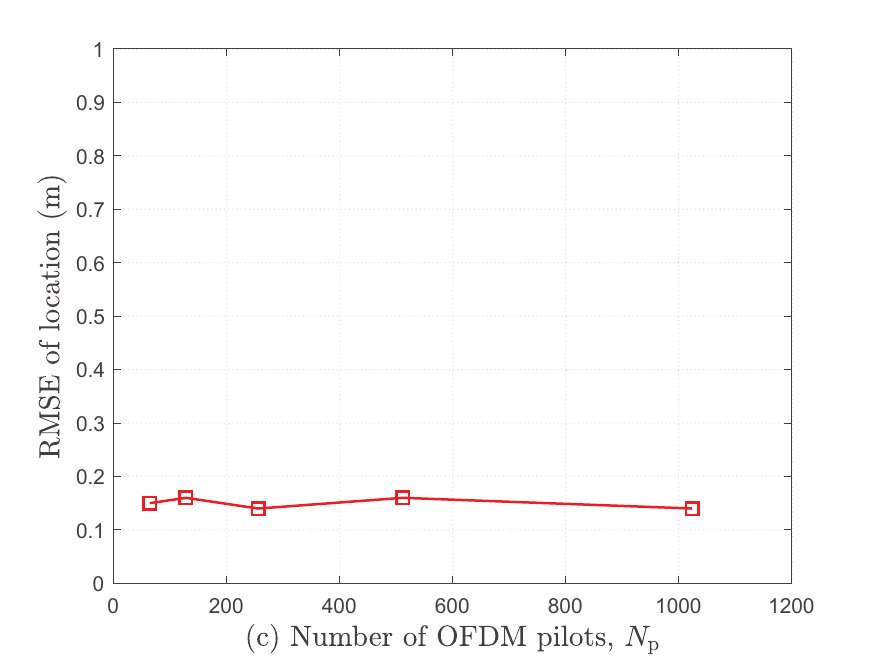}\hspace{-1.2cm}
	}
	\caption{Number of OFDM pilots versus  Top-$K$ accuracy of time delay, Doppler frequency, and RMSE of location.}
	\label{fig:Varying_numpilot_topk}
	
\end{figure*}

\begin{figure*}[!t]  
	\centering
	{
		\hspace{-1.2cm}	\includegraphics[width=0.38\textwidth]{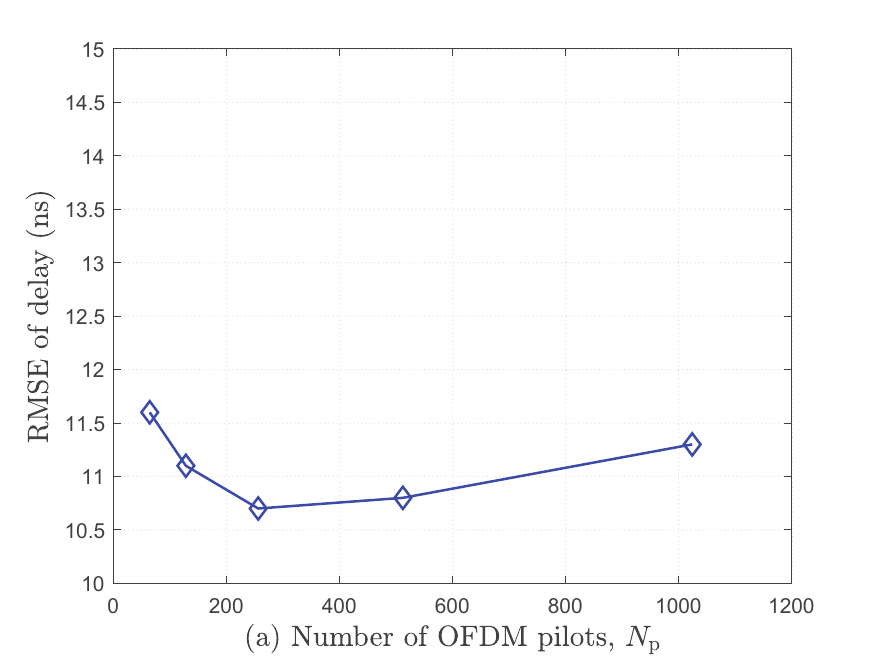}\hspace{-1cm}
	}
	{
		\includegraphics[width=0.38\textwidth]{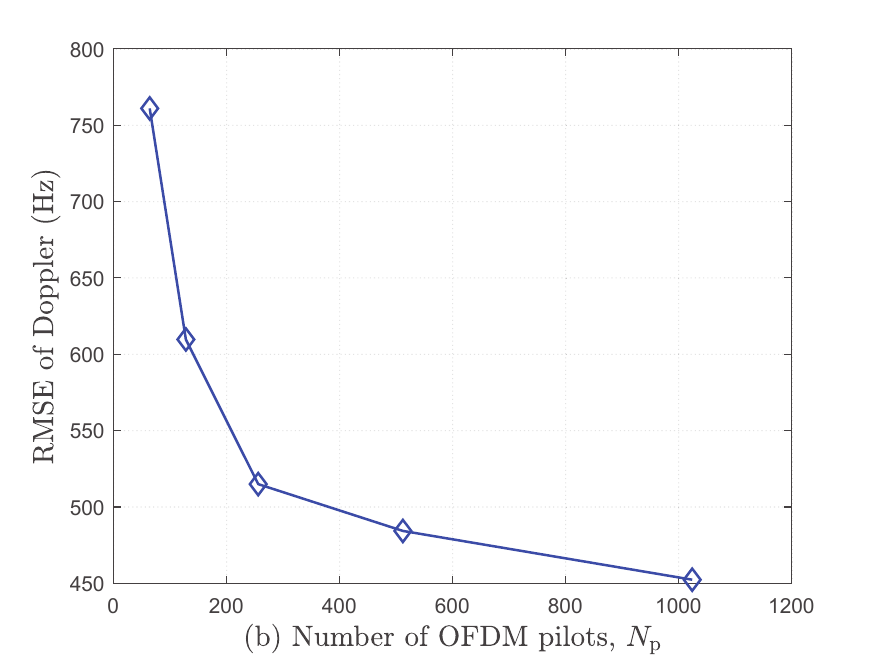}\hspace{-1cm}
	}
	{
		\includegraphics[width=0.38\textwidth]{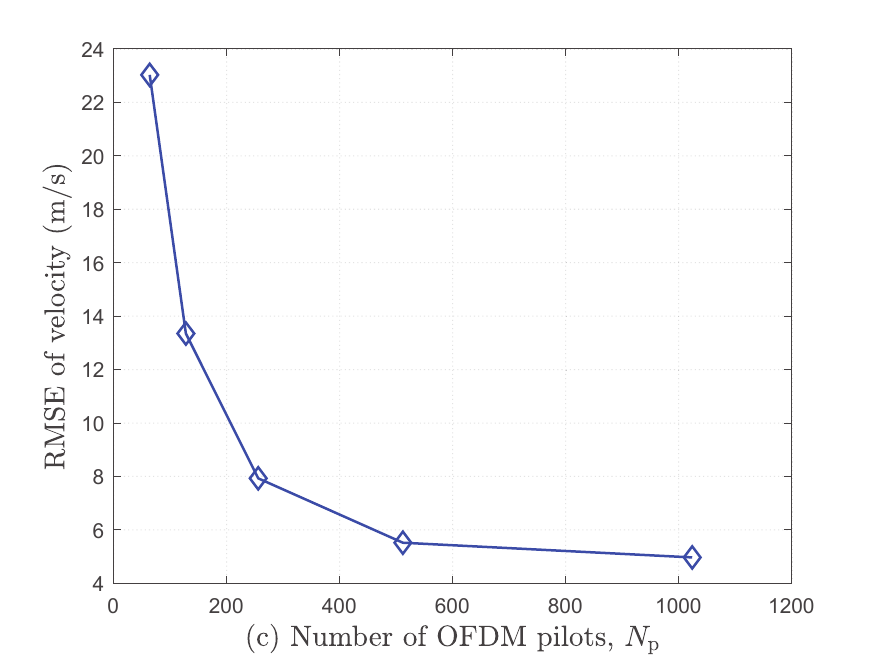}\hspace{-1.2cm}
	}
	\caption{Number of OFDM pilots versus estimated results including RMSE of delay, RMSE of Doppler frequency, and RMSE of velocity.}
	\label{fig:Varying_numpilot_RMSE}
\end{figure*}


Fig.~\ref{fig:Varying_numpilot_topk} investigates the impact of the number of OFDM pilots on the Top-$K$ accuracy of delay and Doppler estimation for $N_{\rm c}=256$ and $\text{SNR} = 20~\text{dB}$, as well as the RMSE of location. As shown in Fig.~\ref{fig:Varying_numpilot_topk}(a), when the number of pilots increases from 32 to 1024, the delay accuracy remains within the range of 87\%–89\%, indicating that the number of pilots has little influence on the delay estimation accuracy. Similarly, Fig.~\ref{fig:Varying_numpilot_topk}(c) shows that the location estimation error stays nearly constant between 0.1 m and 0.2 m, suggesting that the number of pilots has a negligible impact on localization accuracy. In contrast, Fig.~\ref{fig:Varying_numpilot_topk}(b) reveals that the Doppler accuracy decreases as the number of pilots increases. This is because a larger number of pilots expands the output dimension of the Doppler estimation, making the prediction of Doppler bins more challenging. Nevertheless, the increased number of pilots improves Doppler resolution, which ultimately reduces the estimation error of Doppler frequency as clearly shown in Fig.~\ref{fig:Varying_numpilot_RMSE}.

In order to quantitatively evaluate the parameter estimation performance, we compute the RMSE of delay, Doppler frequency, and velocity from the estimated bin indices.  As shown in Fig.~\ref{fig:Varying_numpilot_RMSE}(a), the RMSE of delay fluctuates within the range of 10.5–11.7 ns, which is consistent with the nearly constant delay accuracy observed in Fig.~\ref{fig:Varying_numpilot_topk}(a). On the other hand, Fig.~\ref{fig:Varying_numpilot_RMSE}(b) demonstrates that the RMSE of Doppler frequency significantly decreases from 760 Hz to 450 Hz as the number of pilots increases. This improvement is also reflected in Fig.~\ref{fig:Varying_numpilot_RMSE}(c), where the RMSE of velocity estimation decreases from 23 m/s to 4.8 m/s. These results confirm that, although Doppler accuracy decreases with more pilots due to increased output dimensionality, the overall Doppler and velocity estimation precision benefits from the enhanced resolution.  It is worth noting that although the Doppler frequency estimation error is small, the corresponding velocity estimation error remains relatively large. This is because the considered system is bistatic, and the derived velocity corresponds to the true target velocity rather than the radial velocity. Consequently, even small Doppler errors can translate into larger discrepancies in the estimated true velocity.


\begin{figure*}[!t]  
	\centering
	{
		\hspace{-1.2cm}	\includegraphics[width=0.38\textwidth]{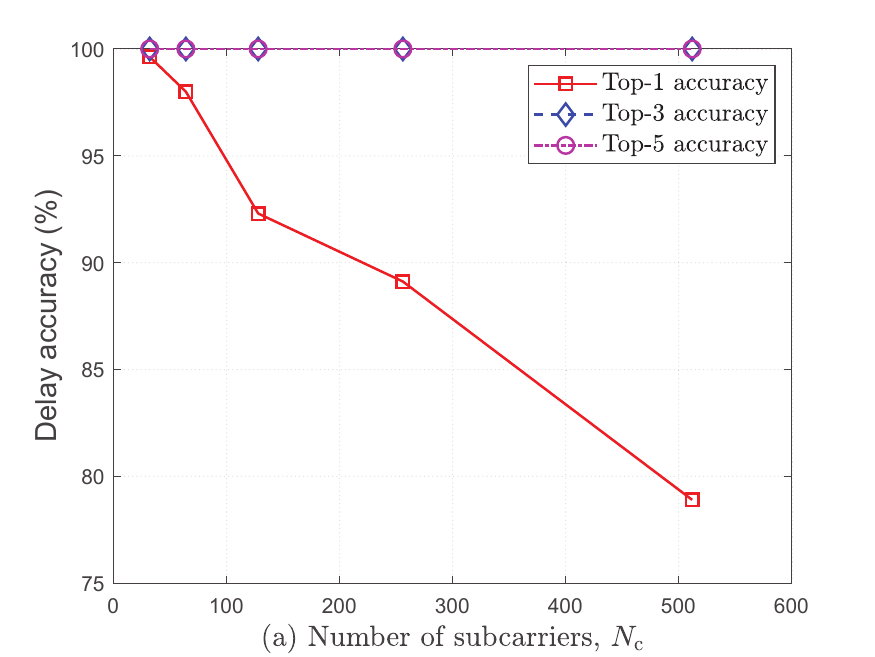}\hspace{-1cm}
	}
	{
		\includegraphics[width=0.38\textwidth]{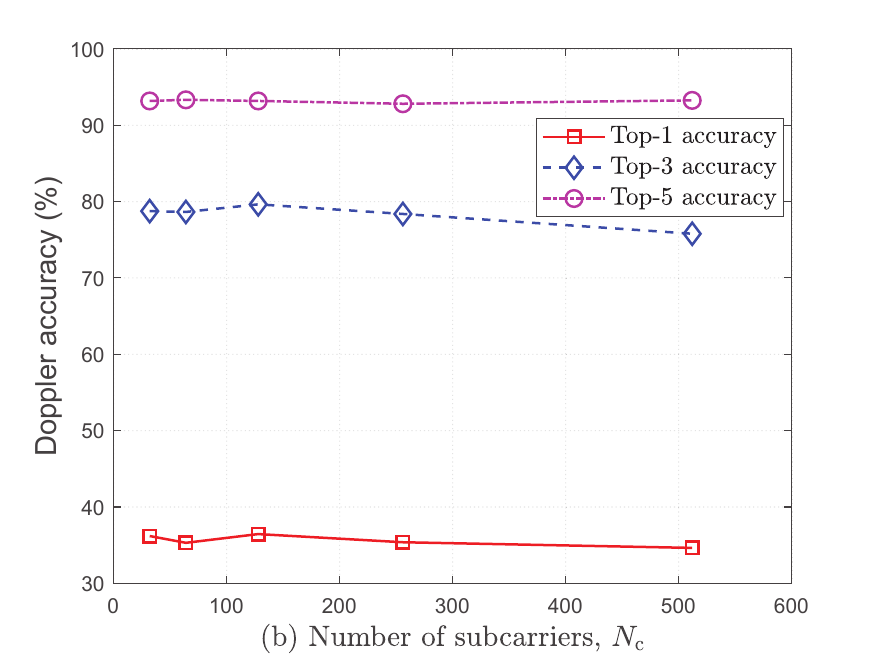}\hspace{-1cm}
	}
	{
		\includegraphics[width=0.38\textwidth]{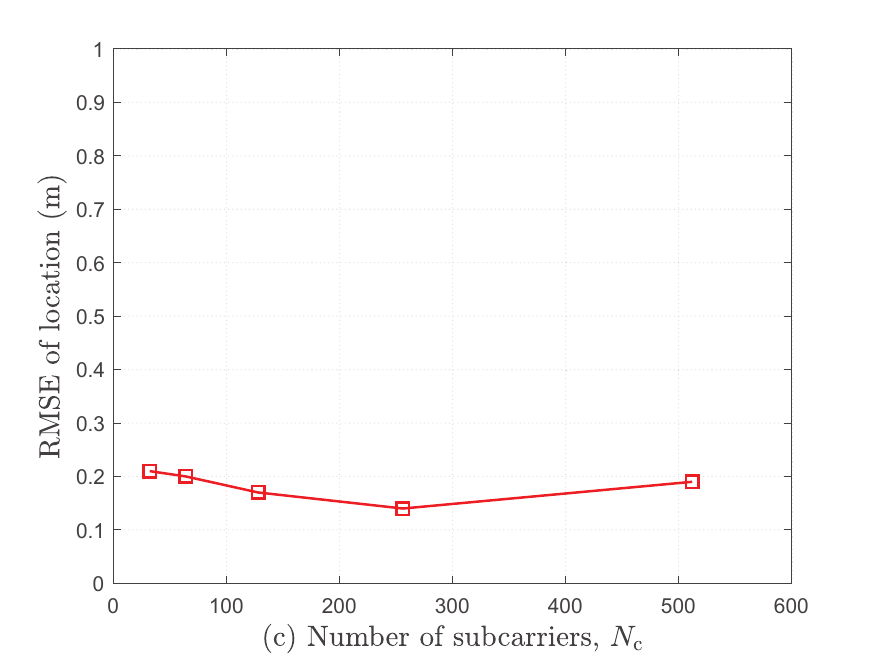}\hspace{-1.2cm}
	}
	\caption{Number of subcarriers versus  Top-$k$  accuracy of delay, Doppler frequency, and RMSE of location.}
	\label{fig:Varying_subcarrier_topk}
	
\end{figure*}

\begin{figure*}[!t]  
	\centering
	{
		\hspace{-1.2cm}	\includegraphics[width=0.38\textwidth]{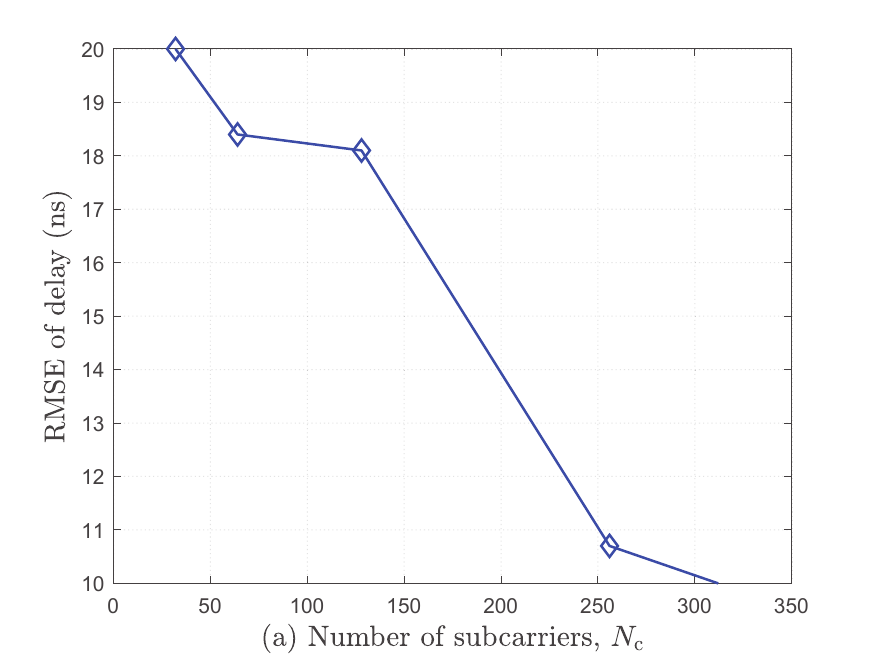}\hspace{-1cm}
	}
	{
		\includegraphics[width=0.38\textwidth]{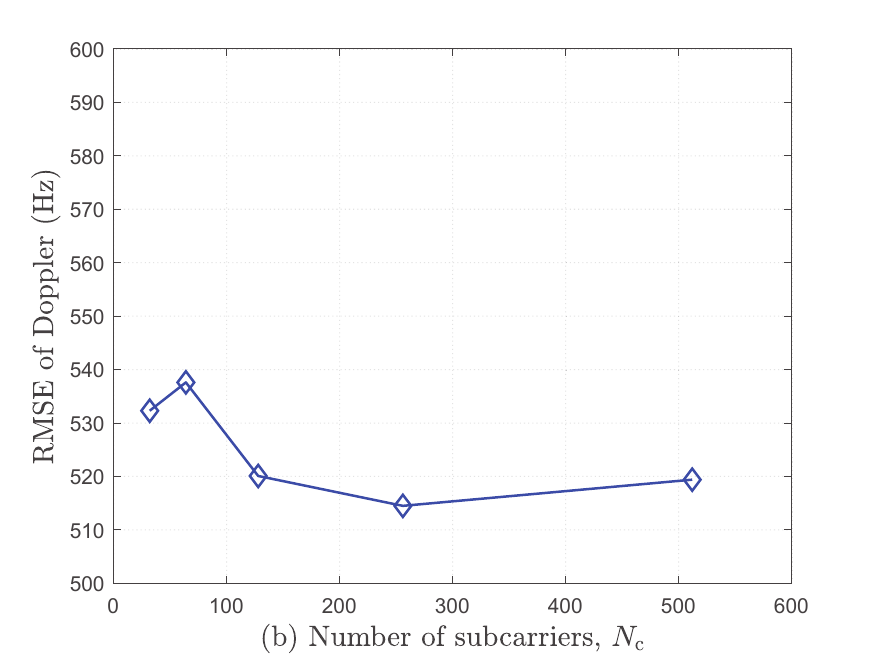}\hspace{-1cm}
	}
	{
		\includegraphics[width=0.38\textwidth]{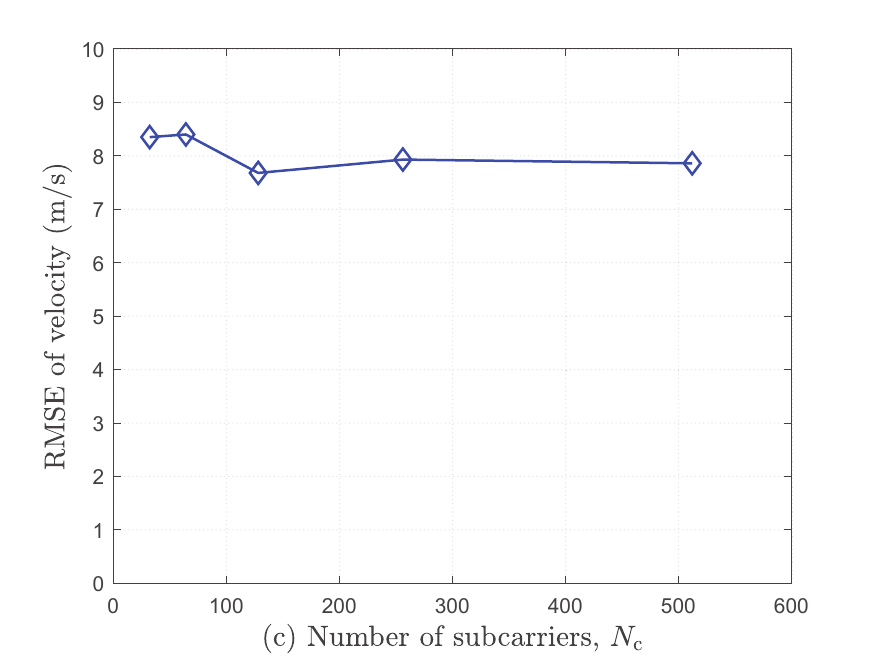}\hspace{-1.2cm}
	}
	\caption{Number of subcarriers versus  RMSE of delay, RMSE of Doppler frequency, and RMSE of velocity.}
	\label{fig:Varying_subcarrier_RMSE}
	
\end{figure*}

Fig.~\ref{fig:Varying_subcarrier_topk} investigates the effect of the number of subcarriers on the Top-$K$ accuracy of delay and Doppler estimation, as well as the RMSE of location, for $N_{\rm p}=256$ and $\text{SNR} = 20~\text{dB}$. As shown in Fig.~\ref{fig:Varying_subcarrier_topk}(a), when the number of subcarriers increases from 32 to 512, the delay accuracy decreases from nearly 100\% to about 78\%. This degradation occurs because a larger number of subcarriers enlarges the output dimension of the delay estimation network, making the classification of delay bins more challenging. Nevertheless, a higher number of subcarriers improves the delay resolution, which in turn reduces the delay estimation error, as will be confirmed in Fig.~\ref{fig:Varying_subcarrier_RMSE}. In contrast, Fig.~\ref{fig:Varying_subcarrier_topk}(b) demonstrates that Doppler accuracy remains almost unchanged across different numbers of  subcarriers, indicating that the number of  subcarriers has little influence on Doppler estimation accuracy. Similarly, Fig.~\ref{fig:Varying_subcarrier_topk}(c) shows that the location estimation error fluctuates between 0.13 m and 0.21 m, suggesting that localization accuracy is unaffected by the number of subcarriers.

To further examine the estimation error, Fig.~\ref{fig:Varying_subcarrier_RMSE} reports the RMSE performance of delay, Doppler frequency, and velocity as a function of the number of subcarriers. As shown in Fig.~\ref{fig:Varying_subcarrier_RMSE}(a), the RMSE of delay decreases significantly from about 20 ns to 10 ns as the number of subcarriers increases. Meanwhile, Fig.~\ref{fig:Varying_subcarrier_RMSE}(b) and Fig.~\ref{fig:Varying_subcarrier_RMSE}(c) reveal that both the RMSE of Doppler frequency and the corresponding velocity estimation remain nearly constant, with velocity errors lying between 7.6 m/s and 8.3 m/s. The reason is that the number of subcarriers mainly determines the delay resolution, whereas the Doppler resolution is determined by the number of pilot OFDM symbols.
These results demonstrate that while increasing the number of subcarriers improves delay resolution, it has a negligible effect on Doppler and location estimation performance.

\begin{figure}[!t]
	\centerline{\includegraphics[width=3.5in]{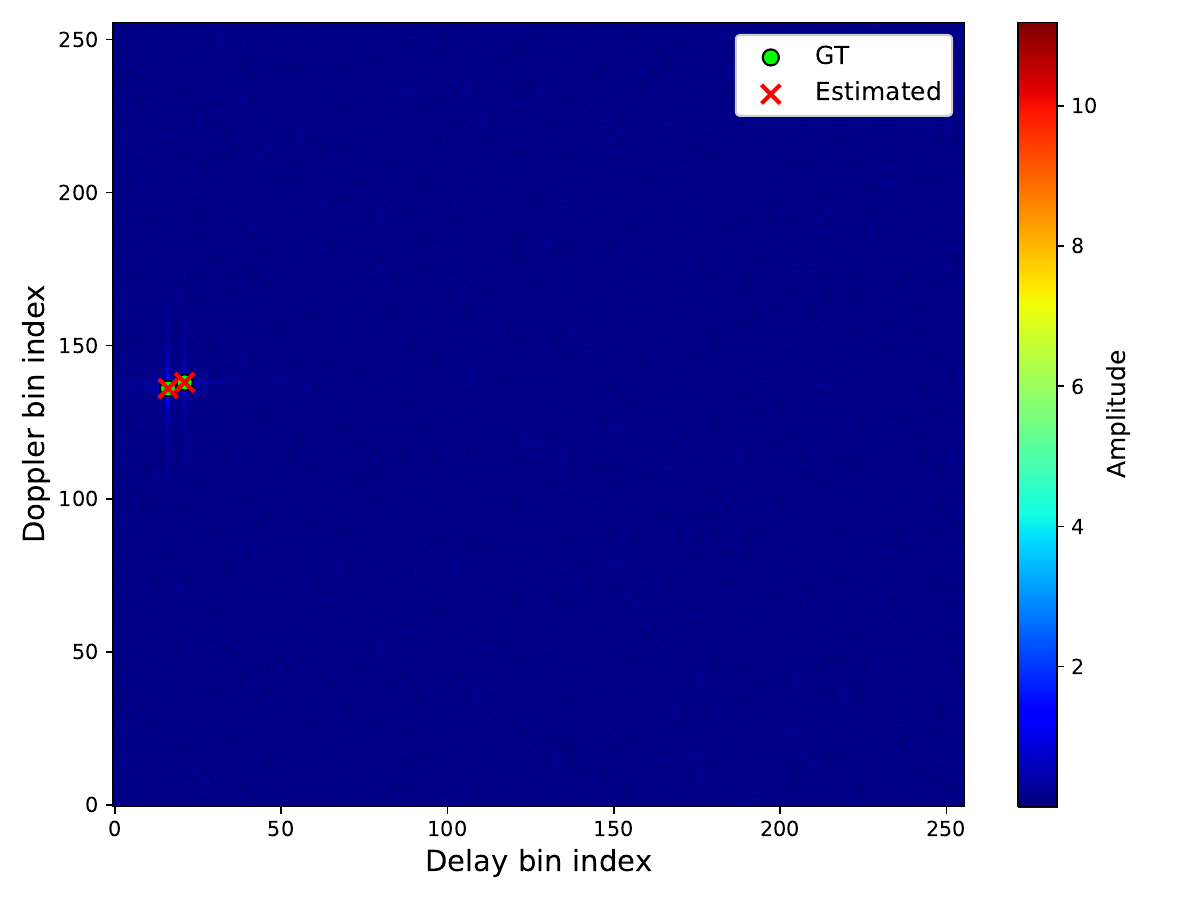}}
	\caption{Visualized DDM achieved by our proposed multi-modal learning scheme.} \label{fig:Visualization_DDM}
	\vspace{-0.3cm}
\end{figure}

Fig.~\ref{fig:Visualization_DDM} illustrates a visualized DDM obtained using the proposed multi-modal learning scheme for $N_{\rm c}=256$, $N_{\rm p}=256$, and $\text{SNR} = 20~\text{dB}$. In the figure, the green circle (GT) denotes the ground-truth position of the vehicle in the delay–Doppler domain, while the red ``$\times$” indicates the corresponding estimate produced by the proposed framework. It can be observed that the estimated positions almost perfectly overlap with the ground-truth values, which confirms the ability of the proposed scheme to accurately recover both delay and Doppler parameters.

\begin{table*}[!t]
	\centering  
	\setlength{\tabcolsep}{5pt}
	\renewcommand{\arraystretch}{1.15}
	\caption{Ablation study results under  $N_{\rm c}=512$, $N_{\rm p}=512$, and  ${\rm SNR}=20~{\rm dB}$.}
	\label{tab:ablation}
	\begin{threeparttable}
		\begin{tabular}{lcccc}
			\Xhline{1pt} 
			\textbf{Scheme} & \textbf{RMSE of location (m)} &   \textbf{RMSE of delay (ns)} &
			 \textbf{RMSE of velocity (m/s)} \\
			\midrule
			
			\textbf{Proposed multi-modal data fusion}                       & \textbf{0.16}  & \textbf{10.82 } & \textbf{5.52}   \\
			w/o prior information       & 0.54  &12.81  & 5.65    \\
			w/o vehicle type                     & 0.16  & 11.24  &5.83   \\
			w/o DDM         & 0.18  & 11.04  & 18.30    \\
			w/o bounding box & 10.69  & 63.9  & 14.29  \\
			\Xhline{1pt}
		\end{tabular}
	\end{threeparttable}\label{Ablation_results}
\end{table*}
Table~\ref{Ablation_results} summarizes the ablation study results under $N_{\rm c}=512$, $N_{\rm p}=512$, and  ${\rm SNR}=20~{\rm dB}$, where different components of the proposed multi-modal data fusion scheme are removed for comparison. As observed, the proposed scheme achieves the best overall performance across all metrics, with the lowest RMSE values of 0.16 m for location,   10.82 ns for delay, and   5.52 m/s for velocity.
When prior information is excluded, the performance of location estimation degrades noticeably, i.e., 0.54 m versus 0.16 m, while delay and velocity estimation also suffer moderate losses. This is because the delay and Doppler are both related to the prior geometric information, which can reduce estimation ambiguity.
In the case without vehicle type, the location accuracy remains nearly identical to the proposed scheme, but both delay and velocity estimates exhibit slightly worse performance, highlighting the role of vehicle category in refining physical parameter estimation. In contrast, when the DDM is excluded, the velocity RMSE increases dramatically to 18.30 m/s, indicating that reliable velocity estimation is almost impossible without DDM information. Nevertheless, location and delay can still be estimated with reasonable accuracy, suggesting that visual features alone can provide meaningful cues for these parameters. Finally, when bounding box information is removed, all three estimation tasks fail, with RMSE values of 10.69 m, 63.9 ns, and 14.29 m/s for location, delay, and velocity, respectively. This result confirms that bounding boxes implicitly encode the vehicle position, and without them, different vehicles cannot be properly associated with their corresponding delay–Doppler pairs, leading to failure in parameter estimation.
\section{Conclusion}
We present a vision-assisted OFDM ISAC framework that fuses a DeepJSCC-reconstructed street-view image, a YOLOv5-extracted per-target feature vector, and the OFDM delay-Doppler map through a learned multi-modal network.
We also designed a novel loss function based on KL divergence with soft-label construction to overcome the instability issues caused by large output dimensions in delay–Doppler classification.   Simulation results on a realistic Blender-based  vehicle testbed  demonstrated that the proposed framework achieves a 16 cm localization RMSE, a 10.8 ns delay RMSE, and a 5.5 m/s velocity RMSE, significantly outperforming conventional ISAC methods. An ablation study confirmed that removing the visual modality degrades localization by nearly two orders of magnitude, showing that vision is a practical solution to mitigate the data-association and resolution limits of single-modality ISAC.
As future work, we plan to extend the framework to nighttime scenarios, where integrating wireless signals with low-light or infrared vision will be crucial for robust sensing and reliable parameter estimation.

\section*{Acknowledgment}
The authors would like to express sincere gratitude to Wei Huang  for his guidance and assistance in generating the Blender-based dataset.

\bibliographystyle{IEEEtran}
\bibliography{Ref_vision_ISAC.bib}

@Article{Bourtsoulatze2019deep,
  Title                    = {Deep Joint Source-Channel Coding for Wireless Image Transmission},
  Author                   = {Bourtsoulatze, Eirina and Burth Kurka, David and G{\"u}nd{\"u}z, Deniz},
  Journal                  = {IEEE Trans. Cogn. Commun. Netw.},
  Year                     = {2019},

  Month                    = {Sept.},
  Number                   = {3},
  Pages                    = {567-579},
  Volume                   = {5}
}

@Article{charan2021vision6G,
  Title                    = {Vision-Aided {6G} Wireless Communications: Blockage Prediction and Proactive Handoff},
  Author                   = {Charan, Gouranga and Alrabeiah, Muhammad and Alkhateeb, Ahmed},
  Journal                  = {IEEE Trans. Veh. Technol.},
  Year                     = {2021},
  Number                   = {10},
  Pages                    = {10193-10208},
  Volume                   = {70}
}

@Article{charan2022multi,
  Title                    = {Multi-modal beam prediction challenge 2022: Towards generalization},
  Author                   = {Charan, Gouranga and Demirhan, Umut and Morais, Jo{\~a}o and Behboodi, Arash and Pezeshki, Hamed and Alkhateeb, Ahmed},
  Journal                  = {arXiv preprint arXiv:2209.07519},
  Year                     = {2022}
}

@InProceedings{chran2022vision,
  Title                    = {Vision-Position Multi-Modal Beam Prediction Using Real Millimeter Wave Datasets},
  Author                   = {Charan, Gouranga and Osman, Tawfik and Hredzak, Andrew and Thawdar, Ngwe and Alkhateeb, Ahmed},
  Booktitle                = {IEEE Wireless Communications and Networking Conference (WCNC)},
  Year                     = {Austin, TX, USA, 2022},
  Pages                    = {2727-2731}
}

@Article{feng2024visionultra,
  Title                    = {Vision-Aided Ultra-Reliable Low-Latency Communications for Smart Factory},
  Author                   = {Feng, Yuan and Gao, Feifei and Tao, Xiaoming and Ma, Shaodan and Poor, H. Vincent},
  Journal                  = {IEEE Trans. Commun.},
  Year                     = {2024},

  Month                    = {Jun.},
  Number                   = {6},
  Pages                    = {3439-3453},
  Volume                   = {72}
}

@Article{gao2025vision,
  Title                    = {Vision-Aided {ISAC} in Low-Altitude Economy Networks via De-Diffused Visual Priors},
  Author                   = {Gao, Yulan and Ye, Ziqiang and Lyu, Zhonghao and Xiao, Ming and Xiao, Yue and Yang, Ping and Manolova, Agata},
  Journal                  = {IEEE Trans. Cogn. Commun. Netw.},
  Year                     = {2026},
  Pages                    = {3831-3845},
  Volume                   = {12}
}

@Article{gonzelez2024integrated,
  Title                    = {The Integrated Sensing and Communication Revolution for {6G}: Vision, Techniques, and Applications},
  Author                   = {Gonz\'{a}lez-Prelcic, Nuria and Furkan Keskin, Musa and Kaltiokallio, Ossi and Valkama, Mikko and Dardari, Davide and Shen, Xiao and Shen, Yuan and Bayraktar, Murat and Wymeersch, Henk},
  Journal                  = {Proc. IEEE},
  Year                     = {2024},

  Month                    = {Jul.},
  Number                   = {7},
  Pages                    = {676-723},
  Volume                   = {112}
}

@Article{huahaocheng2023optimal,
  Title                    = {Optimal Transmit Beamforming for Integrated Sensing and Communication},
  Author                   = {Hua, Haocheng and Xu, Jie and Han, Tony Xiao},
  Journal                  = {IEEE Trans. Veh. Technol.},
  Year                     = {2023},

  Month                    = {Aug.},
  Number                   = {8},
  Pages                    = {10588-10603},
  Volume                   = {72}
}

@Article{hua20243d,
  Title                    = {{3D} Multi-Target Localization via Intelligent Reflecting Surface: Protocol and Analysis},
  Author                   = {Hua, Meng and Chen, Guangji and Meng, Kaitao and Ma, Shaodan and Yuen, Chau and Cheung So, Hing},
  Journal                  = {IEEE Trans. Wireless Commun.},
  Year                     = {2024},

  Month                    = {Nov. },
  Number                   = {11},
  Pages                    = {16527-16543},
  Volume                   = {23}
}

@Article{hua2024integrated,
  Title                    = {Integrated Sensing and Communication: Joint Pilot and Transmission Design},
  Author                   = {Hua, Meng and Wu, Qingqing and Chen, Wen and Jamalipour, Abbas and Wu, Celimuge and Dobre, Octavia A.},
  Journal                  = {IEEE Trans. Wireless Commun.},
  Year                     = {2024},

  Month                    = {Nov.},
  Number                   = {11},
  Pages                    = {16017-16032},
  Volume                   = {23}
}

@Article{hua2023joint,
  Title                    = {Joint Active and Passive Beamforming Design for {IRS}-Aided Radar-Communication},
  Author                   = {Hua, Meng and Wu, Qingqing and He, Chong and Ma, Shaodan and Chen, Wen},
  Journal                  = {IEEE Trans. Wireless Commun.},
  Year                     = {2023},

  Month                    = { Apr.},
  Number                   = {4},
  Pages                    = {2278-2294},
  Volume                   = {22}
}

@InProceedings{huang2024vision,
  Title                    = {Vision Image Aided Near-Field Beam Training for Internet of Vehicle Systems},
  Author                   = {Huang, Wei and Huang, Xueqing and Zhang, Haiyang and Sun, Kunyang and Kai, Caihong and He, Shiwen},
  Booktitle                = {IEEE International Conference on Communications Workshops (ICC Workshops)},
  Year                     = {Denver, CO, USA,2024},
  Pages                    = {390-395}
}

@InProceedings{jiang2022computer,
  Title                    = {Computer Vision Aided Beam Tracking in A Real-World Millimeter Wave Deployment},
  Author                   = {Jiang, Shuaifeng and Alkhateeb, Ahmed},
  Booktitle                = {IEEE Globecom Workshops (GC Wkshps)},
  Year                     = { Rio de Janeiro, Brazil, 2022},
  Pages                    = {142-147}
}

@Article{jiang2024ISACnet,
  Title                    = {{ISAC-NET}: Model-Driven Deep Learning for Integrated Passive Sensing and Communication},
  Author                   = {Jiang, Wangjun and Ma, Dingyou and Wei, Zhiqing and Feng, Zhiyong and Zhang, Ping and Peng, Jinlin},
  Journal                  = {IEEE Trans. Commun.},
  Year                     = {2024},
  Number                   = {8},
  Pages                    = {4692-4707},
  Volume                   = {72}
}

@Article{jiang2025networklevel,
  Title                    = {Network-Level Performance Analysis for Air-Ground Integrated Sensing and Communication},
  Author                   = {Jiang, Yihang and Li, Xiaoyang and Zhu, Guangxu and Han, Kaifeng and Meng, Kaitao and Yang, Wu and Liu, Chenji and Shi, Qingjiang and Zhang, Rui},
  Journal                  = {IEEE Trans. Wireless Commun.},
  Year                     = {2025},

  Month                    = {Aug.},
  Number                   = {8},
  Pages                    = {6931-6946},
  Volume                   = {24}
}

@Misc{yolov5,
  Title                    = {{YOLOv5}: Open Source Neural Network for Object Detection},

  Author                   = {Glenn Jocher and Ultralytics},
  HowPublished             = {\emph{GitHub Repository}},
  Note                     = {Available: https://github.com/ultralytics/yolov5},
  Year                     = {2020}
}

@Article{li2025mimo,
  Title                    = {{MIMO-OFDM ISAC} Waveform Design for Range-{Doppler} Sidelobe Suppression},
  Author                   = {Li, Peishi and Li, Ming and Liu, Rang and Liu, Qian and Lee Swindlehurst, A.},
  Journal                  = {IEEE Trans. Wireless Commun.},
  Year                     = {2025},

  Month                    = {Feb.},
  Number                   = {2},
  Pages                    = {1001-1015},
  Volume                   = {24}
}

@Article{limimo2025,
  Title                    = {{MIMO}-{OFDM} {ISAC} Waveform Design for Range-{Doppler} Sidelobe Suppression},
  Author                   = {Li, Peishi and Li, Ming and Liu, Rang and Liu, Qian and Lee Swindlehurst, A.},
  Journal                  = {IEEE Trans. Wireless Commun.},
  Year                     = {2025},

  Month                    = {Feb.},
  Number                   = {2},
  Pages                    = {1001-1015},
  Volume                   = {24}
}

@Article{lin2024multicamera,
  Title                    = {Multi-Camera Views Based Beam Searching and {BS} Selection With Reduced Training Overhead},
  Author                   = {Lin, Bo and Gao, Feifei and Zhang, Yong and Pan, Chengkang and Liu, Guangyi},
  Journal                  = {IEEE Trans. Commun.},
  Year                     = {2024},

  Month                    = { May },
  Number                   = {5},
  Pages                    = {2793-2805},
  Volume                   = {72}
}

@InProceedings{lin2014microsoft,
  Title                    = {Microsoft coco: Common objects in context},
  Author                   = {Lin, Tsung-Yi and Maire, Michael and Belongie, Serge and Hays, James and Perona, Pietro and Ramanan, Deva and Doll{\'a}r, Piotr and Zitnick, C Lawrence},
  Booktitle                = {European conference on computer vision (ECCV)},
  Year                     = {2014},
  Organization             = {Springer},
  Pages                    = {740--755}
}

@Article{Liu2022survey,
  Title                    = {A Survey on Fundamental Limits of Integrated Sensing and Communication},
  Author                   = {Liu, An and Huang, Zhe and Li, Min and Wan, Yubo and Li, Wenrui and Han, Tony Xiao and Liu, Chenchen and Du, Rui and Tan, Danny Kai Pin and Lu, Jianmin and Shen, Yuan and Colone, Fabiola and Chetty, Kevin},
  Journal                  = {IEEE Commun. Surveys Tuts.},
  Year                     = {2022},

  Month                    = {2nd Quat.},
  Number                   = {2},
  Pages                    = {994-1034},
  Volume                   = {24}
}

@InProceedings{liuvortex2023,
  Title                    = {Vortex Wavefront {FMCW} {ISAC} Model: A Blender-Based Evaluation},
  Author                   = {Liu, Yuan and Long, Wen-Xuan and Chen, Rui and Wu, Linlong and Shankar, M. R. Bhavani},
  Booktitle                = {IEEE 24th International Workshop on Signal Processing Advances in Wireless Communications (SPAWC)},
  Year                     = {Shanghai, China, 2023},
  Pages                    = {431-435}
}

@Article{lu2024deep,
  Title                    = {Deep-Learning-Based Multinode {ISAC 4D} Environmental Reconstruction With Uplink–Downlink Cooperation},
  Author                   = {Lu, Bohao and Wei, Zhiqing and Wu, Huici and Zeng, Xinrui and Wang, Lin and Lu, Xi and Mei, Dongyang and Feng, Zhiyong},
  Journal                  = {IEEE Internet Things J.},
  Year                     = {2024},

  Month                    = { Dec.},
  Number                   = {24},
  Pages                    = {39512-39526},
  Volume                   = {11}
}

@Article{yang2024semantic,
  Title                    = {Semantic-Aware Vision-Assisted Integrated Sensing and Communication: Architecture and Resource Allocation},
  Author                   = {Lu, Yang and Mao, Weihao and Du, Hongyang and Dobre, Octavia A. and Niyato, Dusit and Ding, Zhiguo},
  Journal                  = {IEEE Wireless Commun.},
  Year                     = {2024},

  Month                    = {Jun.},
  Number                   = {3},
  Pages                    = {302-308},
  Volume                   = {31}
}

@Article{meng2025network,
  Title                    = {Network-level ISAC: An Analytical Study of Antenna Topologies Ranging from Massive to Cell-Free {MIMO}},
  Author                   = {Meng, Kaitao and Han, Kawon and Masouros, Christos and Hanzo, Lajos},
  Journal                  = {IEEE Trans. Wireless Commun.},
  Year                     = {2025},

  Month                    = {Dec.},
  Number                   = {12},
  Pages                    = {10003-10018},
  Volume                   = {24}
}

@Article{meng2025cooperative,
  Title                    = {Cooperative {ISAC} Networks: Opportunities and Challenges},
  Author                   = {Meng, Kaitao and Masouros, Christos and Petropulu, Athina P. and Hanzo, Lajos},
  Journal                  = {IEEE Wireless Commun.},
  Year                     = {2025},

  Month                    = {Jun.},
  Number                   = {3},
  Pages                    = {212-219},
  Volume                   = {32}
}

@Article{ren2024fundamental,
  Title                    = {Fundamental {CRB}-Rate Tradeoff in Multi-Antenna ISAC Systems With Information Multicasting and Multi-Target Sensing},
  Author                   = {Ren, Zixiang and Peng, Yunfei and Song, Xianxin and Fang, Yuan and Qiu, Ling and Liu, Liang and Ng, Derrick Wing Kwan and Xu, Jie},
  Journal                  = {IEEE Trans. Wireless Commun.},
  Year                     = {2024},

  Month                    = {Apr.},
  Number                   = {4},
  Pages                    = {3870-3885},
  Volume                   = {23}
}

@InProceedings{sagduyu2024jointsensing,
  Title                    = {Joint Sensing and Task-Oriented Communications with Image and Wireless Data Modalities for Dynamic Spectrum Access},
  Author                   = {Sagduyu, Yalin E. and Erpek, Tugba and Yener, Aylin and Ulukus, Sennur},
  Booktitle                = {IEEE International Symposium on Dynamic Spectrum Access Networks (DySPAN)},
  Year                     = {Washington, DC, USA,2024},
  Pages                    = {57-62}
}

@Article{temiz2025deep,
  Title                    = {Deep-Learning-Based Techniques for Integrated Sensing and Communication Systems: State-of-the-Art, Challenges, and Opportunities},
  Author                   = {Temiz, Murat and Zhang, Yongwei and Fu, Yanwei and Zhang, Chi and Meng, Chenfeng and Kaplan, Orhan and Masouros, Christos},
  Journal                  = {IEEE Open J. Commun. Soc.},
  Year                     = {2025},

  Month                    = {Jul.},
  Pages                    = {5940-5968},
  Volume                   = {6}
}

@InProceedings{wang2025ofdm,
  Title                    = {An {OFDM-ISAC} Scheme Based on Zadoff-{Chu} Sequence for {5G NR} Base-Station},
  Author                   = {Wang, Mao and Yoshii, Kazutoshi and Shimamoto, Shigeru and Funakoshi, Mizuki and Yabuki, Ayumu and Funayoshi, Hideto},
  Booktitle                = { IEEE 22nd Consumer Communications \& Networking Conference (CCNC)},
  Year                     = { Las Vegas, NV, USA, 2025},
  Pages                    = {1-4}
}

@Article{wang2024unified,
  Title                    = {Unified {ISAC} Pareto Boundary Based on Mutual Information and Minimum Mean-Square Error Estimation},
  Author                   = {Wang, Shuaijun and Chen, Li and Zhou, Jing and Chen, Yunfei and Han, Kaifeng and You, Changsheng},
  Journal                  = {IEEE Trans. Commun.},
  Year                     = {2024},

  Month                    = {Nov.},
  Number                   = {11},
  Pages                    = {6783-6795},
  Volume                   = {72}
}

@Article{wen2023vision,
  Title                    = {Vision Aided Environment Semantics Extraction and Its Application in mmWave Beam Selection},
  Author                   = {Wen, Feiyang and Xu, Weihua and Gao, Feifei and Pan, Chengkang and Liu, Guangyi},
  Journal                  = {IEEE Commun. Lett.},
  Year                     = {2023},

  Month                    = {Jul.},
  Number                   = {7},
  Pages                    = {1894-1898},
  Volume                   = {27}
}

@Article{wu2025low,
  Title                    = {Low-Complexity Minimum BER Precoder Design for {ISAC} Systems: A Delay-Doppler Perspective},
  Author                   = {Wu, Jun and Yuan, Weijie and Wei, Zhiqiang and Zhang, Kecheng and Liu, Fan and Wing Kwan Ng, Derrick},
  Journal                  = {IEEE Trans. Wireless Commun.},
  Year                     = {2025},

  Month                    = { Feb.},
  Number                   = {2},
  Pages                    = {1526-1540},
  Volume                   = {24}
}

@Article{xiong2023onthefundamental,
  Title                    = {On the Fundamental Tradeoff of Integrated Sensing and Communications Under Gaussian Channels},
  Author                   = {Xiong, Yifeng and Liu, Fan and Cui, Yuanhao and Yuan, Weijie and Han, Tony Xiao and Caire, Giuseppe},
  Journal                  = {IEEE Trans. Inf. Theory},
  Year                     = {2023},

  Month                    = {Sept.},
  Number                   = {9},
  Pages                    = {5723-5751},
  Volume                   = {69}
}

@Article{xu2023computer,
  Title                    = {Computer Vision Aided mmWave Beam Alignment in {V2X} Communications},
  Author                   = {Xu, Weihua and Gao, Feifei and Tao, Xiaoming and Zhang, Jianhua and Alkhateeb, Ahmed},
  Journal                  = {IEEE Trans. Wireless Commun.},
  Year                     = {2023},

  Month                    = {Apr.},
  Number                   = {4},
  Pages                    = {2699-2714},
  Volume                   = {22}
}

@Article{xu2025computerscheudling,
  Title                    = {Computer Vision-Based Link Scheduling in mmWave Multi-Hop {V2X} Communications},
  Author                   = {Xu, Weihua and Zhao, Chuanbin and Gao, Feifei and Xing, Ling and Qian, Jing and Wang, Hao},
  Journal                  = {IEEE Trans. Commun.},
  Year                     = {2025},

  Month                    = {Aug.},
  Number                   = {8},
  Pages                    = {5621-5634},
  Volume                   = {73}
}

@Article{yang2024sensing,
  Title                    = {Sensing Aided Uplink Transmission in {OTFS} {ISAC} With Joint Parameter Association, Channel Estimation and Signal Detection},
  Author                   = {Yang, Xi and Li, Hang and Guo, Qinghua and Zhang, J. Andrew and Huang, Xiaojing and Cheng, Zhiqun},
  Journal                  = {IEEE Trans. Veh. Technol.},
  Year                     = {2024},
  Number                   = {6},
  Pages                    = {9109-9114},
  Volume                   = {73}
}

@Article{zecchin2022lidar,
  Title                    = {{LIDAR} and Position-Aided mmWave Beam Selection With Non-Local {CNNs} and Curriculum Training},
  Author                   = {Zecchin, Matteo and Mashhadi, Mahdi Boloursaz and Jankowski, Mikolaj and G{\"u}nd{\"u}z, Deniz and Kountouris, Marios and Gesbert, David},
  Journal                  = {IEEE Trans. Veh. Technol.},
  Year                     = {2022},

  Month                    = {Mar.},
  Number                   = {3},
  Pages                    = {2979-2990},
  Volume                   = {71}
}

@InProceedings{zhang2025multimodal,
  Title                    = {Multimodal deep learning-empowered beam prediction in future {THz ISAC} systems},
  Author                   = {Zhang, Kai and Yu, Wentao and He, Hengtao and Song, Shenghui and Zhang, Jun and Letaief, Khaled B},
  Booktitle                = { IEEE 36th International Symposium on Personal, Indoor and Mobile Radio Communications (PIMRC)},
  Year                     = {2025}
}

@InProceedings{zhang2022vision,
  Title                    = {Vision Aided Beam Tracking and Frequency Handoff for mmWave Communications},
  Author                   = {Zhang, Tengyu and Liu, Jun and Gao, Feifei},
  Booktitle                = { IEEE Conference on Computer Communications Workshops (INFOCOM)},
  Year                     = {2022},
  Pages                    = {1-2}
}

@Article{zhao2024joint,
  Title                    = {Joint Beamforming and Scheduling for Integrated Sensing and Communication Systems in {URLLC}: A {POMDP} Approach},
  Author                   = {Zhao, Xiaoyu and Angela Zhang, Ying-Jun},
  Journal                  = {IEEE Trans. Commun.},
  Year                     = {2024},

  Month                    = {Oct.},
  Number                   = {10},
  Pages                    = {6145-6161},
  Volume                   = {72}
}

\end{document}